%% file: main.tex
\PassOptionsToPackage{table}{xcolor}
\documentclass{article} 
\usepackage{iclr2027_conference,times}

\input{math_commands.tex}

\usepackage{hyperref}
\usepackage{url}
\usepackage[T1]{fontenc}
\usepackage{amsmath}
\usepackage{amssymb}
\usepackage{amsfonts}

\usepackage{graphicx}
\usepackage{float}
\usepackage{multirow}
\usepackage{booktabs}
\usepackage{array}
\usepackage{tabularx}
\usepackage{subcaption}
\usepackage{placeins}
\usepackage{xcolor}
\usepackage{colortbl}

\definecolor{pglhighlight}{RGB}{235,243,250}

\usepackage{xcolor}
\usepackage{tcolorbox}
\usepackage{soul}

\usepackage{nicefrac}
\usepackage{microtype}
\usepackage{marvosym}
\usepackage{siunitx}

\usepackage{pifont}

\definecolor{Highlight}{rgb}{0.21,0.49,0.74}
\definecolor{tableblue}{RGB}{232,241,249}
\definecolor{tableyellow}{RGB}{255,249,215}

\newcommand{\cmark}{\checkmark}
\newcommand{\xmark}{$\times$}

\newlength\savewidth

\newcolumntype{x}[1]{>{\centering\arraybackslash}p{#1pt}}
\newcolumntype{y}[1]{>{\raggedright\arraybackslash}p{#1pt}}
\newcolumntype{z}[1]{>{\raggedleft\arraybackslash}p{#1pt}}
\newcolumntype{Y}{>{\centering\arraybackslash}X}

\usepackage{booktabs}
\usepackage{tabularx}
\usepackage{wrapfig}
\usepackage[table]{xcolor}
\usepackage{caption}
\usepackage{makecell}

\definecolor{tableblue}{RGB}{225,239,247}
\definecolor{tableyellow}{RGB}{255,248,214}

\iclrfinalcopy

\title{PGL-3D: Towards Progressive Geometric Learning for 3D Visual Query Localization}

\author{%
\parbox[t]{\dimexpr\textwidth-2\tabcolsep\relax}{%
\centering
\bfseries
Liang Peng$^{1}$\thanks{%
Equal contribution.\quad
$^{\dagger}$Corresponding author.%
}
\quad
Shizhuo Mu$^{1,*}$ \quad
Bohan Tan$^{1}$ \quad
Wenyuan Wang$^{1}$ \quad
Chen Zhao$^{1}$
\\[2pt]
Xingping Dong$^{1,\dagger}$ \quad
Heng Fan$^{2}$ \quad
Libo Zhang$^{3}$ \quad
Bo Du$^{1}$
\\[4pt]
\normalfont\footnotesize
$^{1}$School of Computer Science, National Engineering Research Center for Multimedia Software,
\\[-1pt]
Institute of Artificial Intelligence, Hubei Key Laboratory of Multimedia and Network
\\[-1pt]
Communication Engineering, Wuhan University
\\[2pt]
$^{2}$University of North Texas
\qquad
$^{3}$Institute of Software, Chinese Academy of Sciences
}}

\begin{document}
\maketitle
\fancyhead{}
\renewcommand{\headrulewidth}{0pt}

\raggedbottom


\begin{abstract}
3D Visual Query Localization (3DVQL) retrieves the latest contiguous 
occurrence of a queried object in an RGB--point-cloud sequence and predicts 
a 9-DoF cuboid for every response frame. 
The query is captured independently of the search sequence, so its annotated 
pose may differ from how the object appears in the search frames. 
The benchmark baseline predicts cuboids after feature modeling, leaving their 
geometry unused for subsequent feature refinement.
We investigate whether complete intermediate cuboids can improve query and proposal representations before final decoding. 
To this end, we introduce the \textbf{P}rogressive \textbf{G}eometric \textbf{L}earning for \textbf{3DVQL} (\textbf{PGL-3D}), a predict--select--refine--re-predict framework that 
uses complete intermediate cuboids to guide the aggregation of search 
evidence and update both query and proposal representations.
 A shared head first predicts a complete cuboid for every proposal. 
 Query--Tube--Memory (QTM) then selects reference observations by combining proposal 
 association, cuboid quality, frame response, and target absence, because association 
 confidence alone establishes neither target presence nor geometric accuracy. 
 The center, size, and orientation of each selected cuboid define soft pooling weights 
 over the query-conditioned proposal features of its frame. The pooled memory updates 
 both the query and the proposal representations, and the head re-predicts from the 
 updated features. A training-only objective, \textbf{ST-D9O}, supervises the complete 
 cuboid geometry at every stage by adding boundary, signed-distance, and soft-overlap terms to parameter regression.
 PGL-3D achieves a mean stAP of $0.270 \pm 0.004$ on 3DVQL, compared with $0.044$ reported 
 for LaF. Component ablations support the benefits of geometry-guided feature updates beyond 
 intermediate supervision, while stage-wise analyses show improved cuboid accuracy. 
 Replacing the geometry objective in our PROT3D reproduction with ST-D9O further improves 
 mAO on GSOT3D from $21.63\%$ to $25.78\%$. Our code and models will be released.
 
\end{abstract}

\vspace{-5pt}
\section{Introduction}
\begin{figure}[t]
  \centering
  \includegraphics[width=\textwidth]{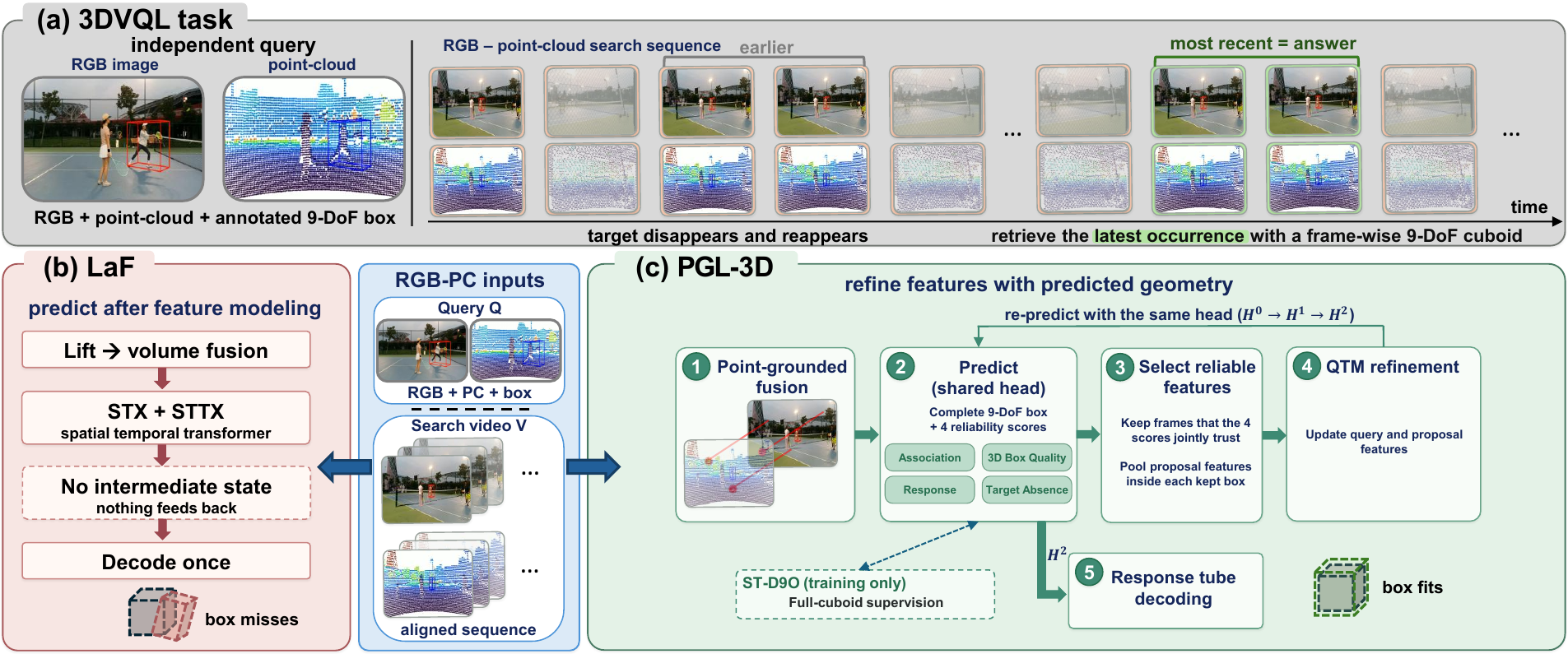}
  \caption{Illustration of the 3DVQL task in (a) and comparison between
LaF in (b) and our PGL-3D in (c). LaF predicts cuboids after feature modeling, whereas PGL-3D
uses intermediate cuboids and reliability estimates to refine
query and proposal features before re-prediction.
ST-D9O provides full-cuboid supervision during training only.}  
  \label{fig:demo_comparison}
\end{figure}

 \vspace{-5pt}
Visual Query Localization (VQL) retrieves an object's latest occurrence
in a video and localizes it with frame-wise 2D boxes~\citep{jiang2023vqloc,grauman2022ego4d,xu2023where,xu2022negative,prvql}. 
Its 3D counterpart, 3DVQL~\citep{laf}, takes an independently captured
RGB--point-cloud query annotated with a cuboid and an RGB--point-cloud
search sequence, and returns the latest contiguous response interval
with a full 9-DoF cuboid in each frame (see Fig.~\ref{fig:demo_comparison}(a)). 
Accurate retrieval requires recognizing the target across
prolonged absence and repeated appearances, while recovering the
target's 3D center, dimensions, and full orientation from partial
observations.

 

LaF, the 3DVQL benchmark baseline, predicts target presence and frame-wise
cuboids after spatio-temporal feature modeling~\citep{laf}, leaving
these estimates unavailable to guide subsequent feature updates (Fig.~\ref{fig:demo_comparison}(b)). 
In 2D VQL, PRVQL already uses intermediate predictions to refine query
and video features~\citep{prvql}. 
Here, we focus on the role of complete intermediate geometry in 3DVQL. 
This raises the central question of 
our work: \emph{can intermediate response and 9-DoF predictions refine 
the query and proposal features before the final response tube is decoded?} 
 
Search observations can complement the independent query, but unreliable
references may reinforce prediction errors.
A proposal can strongly match the query while misestimating the
target's spatial extent or orientation, and high association scores can 
also occur in target-absent frames.
Using these predictions indiscriminately as references risks
reinforcing errors. Therefore, reference selection must consider 
object association, frame-level presence, and cuboid accuracy.
 
We introduce \textbf{PGL-3D}, a predict--select--refine--re-predict framework (Fig.~\ref{fig:demo_comparison}(c)). 
Point-grounded RGB--point-cloud features initialize query-conditioned
proposals, from which a shared head predicts complete cuboids and
separate association, cuboid quality, response, and target-absence
signals.
\textbf{Query--Tube--Memory (QTM)} refinement combines these signals
to select reference observations.
Selected cuboids define box-normalized soft pooling of
query-conditioned proposal features into \emph{Box-Local Memory}.
This memory updates the query tokens and the proposal features, and the
shared head then re-predicts cuboids from the updated representations.
Intermediate geometry thus controls how search evidence is aggregated,
in addition to serving as a prediction target.
After two QTM rounds, final response-tube decoding retrieves the
latest occurrence.
 
Because intermediate cuboids determine how evidence is pooled, we
also supervise their complete geometry.
Separate parameter losses do not directly measure the boundary and
overlap discrepancies induced jointly by center, size, and rotation
errors.
We introduce the \textbf{Spatio-Temporal Differentiable 9-DoF Operator
(ST-D9O)}, which derives cuboid boundaries, signed-distance fields,
and soft occupancy from predicted and target boxes.
Its training objective complements parameter regression with
discrepancies in these geometric representations across prediction
stages, without adding inference-time computation.
 
 \vspace{-10pt}
\paragraph{Contributions.}
\ding{171}~We present \textbf{PGL-3D}, a point-aligned RGB–3D framework that 
predicts complete 9-DoF boxes at multiple stages and refines features with 
its own trusted predictions;\ding{170}~We introduce \textbf{QTM}, which selects 
reliable references with four decoupled signals, constructs Box-Local Memory, 
and updates the query and proposal features before the next prediction 
stage and the final response-tube decoding;\ding{168}~We propose \textbf{ST-D9O}, a training-only 
objective that directly supervises complete oriented cuboid geometry;\ding{169}~Extensive 
experiments establish substantially improved 3DVQL performance and demonstrate 
that ST-D9O transfers to other 3D tracking architectures.
 
\vspace{-5pt}
\section{Related Work}
\label{sec:related_work}
\vspace{-5pt}
 
\paragraph{Visual Query Localization in 2D and 3D.}
Visual Query Localization (VQL) retrieves an object's latest occurrence
and predicts its response interval and frame-wise 2D boxes~\citep{grauman2022ego4d}. 
Subsequent methods study negative-frame modeling, proposal-set matching,
and joint response-and-box prediction~\citep{xu2022negative,xu2023where,jiang2023vqloc}. 
The related Ego4D 
Visual Queries 3D Localization (VQ3D) setting extends the answer to 3D.
However, it asks only for the object center as a displacement vector from the 
query-frame camera, with no box size or orientation~\citep{grauman2022ego4d}.
For this setting, EgoLoc 
estimates camera poses, backprojects 2D response-track detections with depth, and 
aggregates the multi-view estimates weighted by detection confidence~\citep{egoloc}. 
In contrast, 3D Visual Query Localization (3DVQL) introduces retrieval of the 
latest response interval with frame-wise 9-DoF cuboids from RGB--point-cloud 
sequences, together with the LaF baseline~\citep{laf}. We follow this full-cuboid retrieval setting.
 
\vspace{-8pt}
\paragraph{Prediction- and quality-guided refinement.}
PRVQL uses high-confidence 2D regions and attention-derived spatial
cues to refine query and video features~\citep{prvql}.
In 3D tracking, PTTR uses coarse predictions to guide local feature
pooling~\citep{pttr}, while PROT3D refines search features using
intermediate centers, targetness masks, and proposal scores before
final 9-DoF prediction~\citep{gsot3d}.
In 3DVQL, the query is an independent observation, and the target 
must be retrieved at its latest occurrence without a first-frame initialization~\citep{laf}.
IoU-Net estimates localization quality~\citep{jiang2018acquisition},
while Generalized Focal Loss and VarifocalNet incorporate quality into classification
targets~\citep{li2022generalized,zhang2021varifocalnet}.
PGL-3D uses association, cuboid quality, response, and absence to select
references whose complete geometry defines soft pooling into
Box-Local Memory.
The memory updates both query and proposal features, coupling
quality-guided reference selection with full-cuboid-conditioned
feature feedback.
 
\vspace{-8pt}
\paragraph{Optimization of oriented 3D boxes.}
Most 3D detection and tracking methods supervise box center, dimensions, and
rotation with separate losses, which do not directly measure full-box geometry. 
The continuous 6D rotation representation removes discontinuities in the rotation output but defines no box-level objective~\citep{zhou6drotation}.
 KFIoU and 
RDIoU introduce overlap-related objectives whose published 3D formulations 
parameterize rotation by a single yaw angle~\citep{kfiou,rdiou}.
MGIoU projects the vertices of two convex shapes onto their unique face
normals, computes a one-dimensional GIoU on each axis, and averages the
results into a single differentiable overlap loss~\citep{mgiou}.
VSRD uses a cuboid signed-distance function for rendering-based
optimization with weak 2D supervision~\citep{vsrd}.
Bounding Box Disparity compares 3D boxes with full rotational freedom, but it is
proposed as an evaluation metric rather than a training objective~\citep{boundingboxdisparity}. 
The 3DVQL baseline notes that no backpropagatable 9-DoF IoU operator or
combined loss is available~\citep{laf}. ST-D9O provides a differentiable
surrogate for this setting. It combines parameter, boundary,
signed-distance, and soft-overlap terms, marginalizes local-axis flips in
the rotation and boundary terms, and is applied at every prediction stage
with dense and temporal supervision (Sec.~\ref{sec:gst_training}). An
overlap loss such as MGIoU corresponds to the soft-overlap term of ST-D9O
alone; Sec.~\ref{sec:ablation_study} compares the two.

\section{Method}
\label{sec:method}
\begin{figure}[t]
    \centering
    \includegraphics[width=\linewidth]{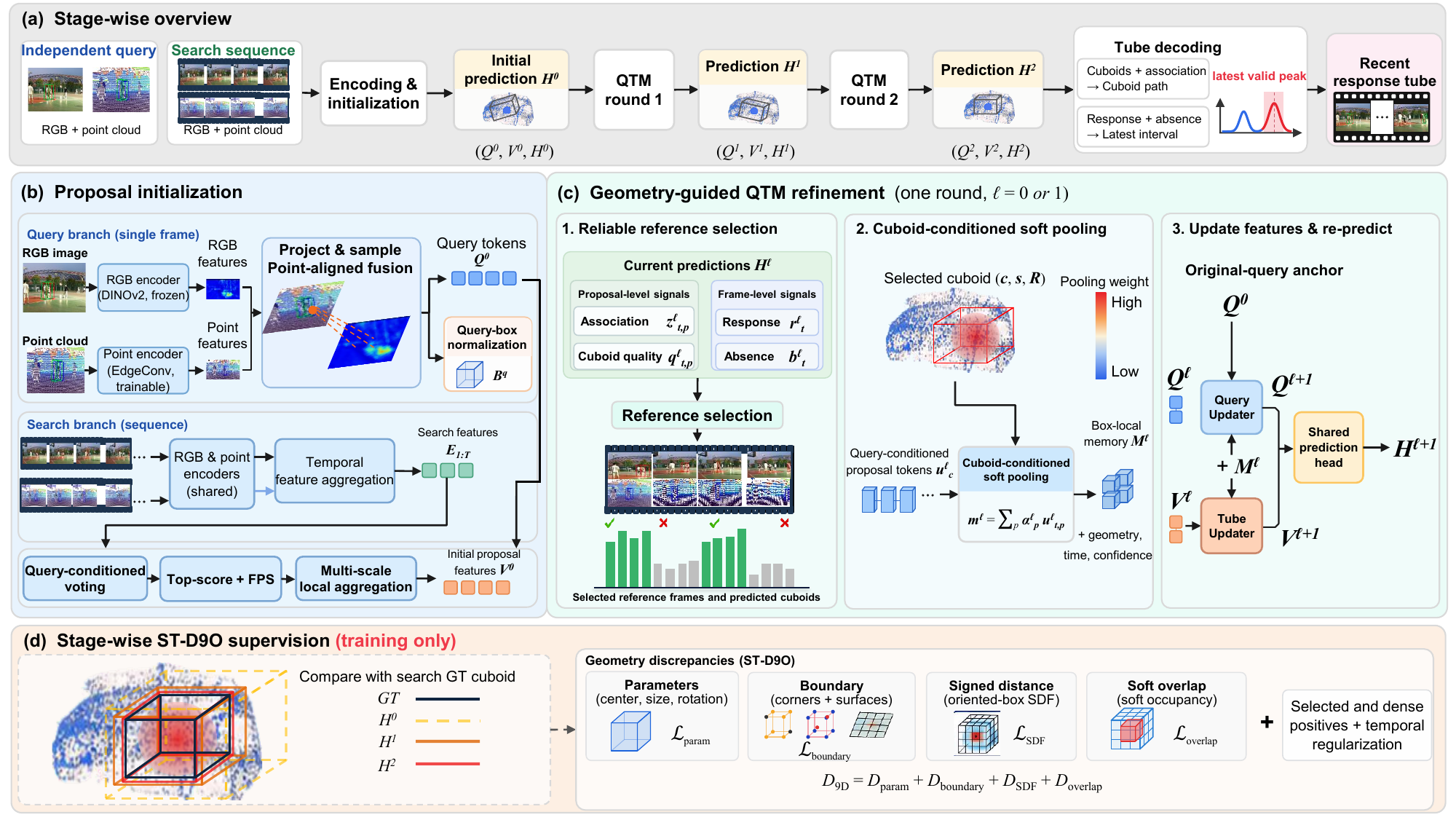}
    \caption{
\textbf{Overview of PGL-3D.}
\textbf{(a)} A shared prediction head produces $H^0$, $H^1$,
and $H^2$, with two Query--Tube--Memory (QTM) refinement
rounds between stages. Final decoding combines a cuboid path
and the latest response interval into the output response tube.
\textbf{(b)} Point-aligned RGB--3D fusion and search-side
temporal aggregation provide features for query-conditioned
voting, proposal sampling, and multi-scale local aggregation.
\textbf{(c)} QTM combines proposal association and cuboid
quality with frame response and absence to select reference
observations. Each selected cuboid's center, size, and
orientation determine soft pooling weights over all
query-conditioned proposal tokens in its reference frame.
The resulting box-local memory updates query and proposal
features through separate updaters before shared-head
re-prediction; proposal centers remain fixed.
\textbf{(d)} Training-only ST-D9O complements parameter
regression with boundary, signed-distance, and soft-overlap
supervision at all three prediction stages. The geometry
discrepancy is applied to selected and dense positive
predictions, with additional temporal regularization.
}
    \label{fig:gst9d_overview}
\end{figure}

 \vspace{-5pt}
\paragraph{Overview.}
PGL-3D uses intermediate 9-DoF predictions to refine query and proposal
features before decoding the final response tube. As shown in
Fig.~\ref{fig:gst9d_overview}, multimodal encoding extracts point-grounded
query and search features (Sec.~\ref{sec:gst_encoding}). We initialize
proposals and predict their cuboids and reliability signals
(Sec.~\ref{sec:gst_prediction}). Query--Tube--Memory (QTM) uses these
predictions to select reference observations, construct box-local memory,
and update both feature sets (Sec.~\ref{sec:gst_qtm}). After two QTM
rounds, the final predictions are decoded into the latest response tube
(Sec.~\ref{sec:gst_decoding}). ST-D9O supervises complete cuboid geometry
during training (Sec.~\ref{sec:gst_training}).
 
  \vspace{-5pt}
\subsection{Multimodal Encoding}
\label{sec:gst_encoding}
 
Given an independent query
$\mathcal Q=(I^q,P^q,\mathcal C^q,B^q)$ and a search sequence
$\mathcal V=\{(I_t,P_t,\mathcal C_t)\}_{t=1}^{T}$, we extract query tokens
and point-level search features. Here, $I$, $P$, and $\mathcal C$ denote
the RGB image, point cloud, and camera calibration. The annotated query
cuboid $B^q=(\mathbf c^q,\mathbf s^q,\mathbf R^q)$ is available during both
training and inference. A cuboid has a center in $\mathbb R^3$, three
positive side lengths, and a rotation in $\mathrm{SO}(3)$.
 
\paragraph{Point-grounded feature extraction.}
A frozen DINOv2 encoder~\citep{dinov2} extracts RGB features, while a
cascaded EdgeConv encoder~\citep{dgcnn} extracts point-cloud features.
Using camera calibration, we project each retained 3D point onto its RGB
feature map and bilinearly sample the corresponding descriptor. A gated
residual combines this descriptor with the point feature, following the
point-painting principle~\citep{pointpainting}. Query geometry is 
expressed in the local coordinates of the annotated cuboid and normalized 
by its half-lengths, whereas RGB sampling uses the original calibrated 
coordinates. Search geometry remains in the camera coordinates of each 
frame; no query-to-search or inter-frame registration is estimated.
 
\vspace{-5pt}
\paragraph{Query and search representations.}
The query features are summarized into $K_q$ tokens. Temporal Feature 
Aggregation enriches search features with nearby-frame 
evidence using feature differences, relative coordinates, time offsets, and
previous-frame displacements, all measured in the camera coordinates of
each frame. The module outputs
\begin{equation}
(Q^0,E_{1:T})=\operatorname{MultimodalEncoder}(\mathcal Q,\mathcal V),
\qquad
E_t=\{(\mathbf x_{t,i},\widetilde{\mathbf e}_{t,i})\}_{i=1}^{M},
\label{eq:gst-encoding}
\end{equation}
where $Q^0\in\mathbb R^{K_q\times D}$ holds the query tokens;
$\mathbf x_{t,i}$ and $\widetilde{\mathbf e}_{t,i}\in\mathbb R^D$ are the
coordinate and the enriched feature of the $i$-th of $M$ retained points in
frame $t$; and $D$ is the feature dimension. Encoding and aggregation are
detailed in
Appendices~\ref{app:evidence_details}--\ref{app:temporal_point_memory}.
 
\vspace{-5pt}
\subsection{3D Feature Learning and Prediction}
\label{sec:gst_prediction}
\vspace{-5pt}
We organize search features into query-conditioned proposals and predict
complete cuboids to provide geometric evidence for QTM. The prediction
head is shared across the initial stage and both refinement stages.
 
\paragraph{Proposal initialization.}
Conditioned on the pooled query, each search point predicts a center vote
and a validity logit~\citep{p2b}. Half of the proposal centers are selected
from the highest-scoring votes and half by farthest-point sampling,
balancing query relevance and spatial coverage. Multi-scale aggregation
collects current-frame point evidence around each center, with neighborhood
radii scaled by $\bar s^q=(s_x^qs_y^qs_z^q)^{1/3}$, the geometric mean of the query side lengths. This gives
\begin{equation}
\bigl(V^0,\{\mathbf p_{t,p}\}_{t,p}\bigr)
=\operatorname{ProposalInit}(E_{1:T},Q^0;\bar s^q),
\label{eq:gst-proposals}
\end{equation}
where $V^0\in\mathbb R^{T\times P\times D}$ contains the initial proposal
features and $\mathbf p_{t,p}$ is the center of proposal $p$ in frame $t$.
These centers remain fixed during QTM.
 
\paragraph{Shared prediction head.}
At stage $\ell\in\{0,1,2\}$, the head combines the current query and
proposal features to predict
\begin{equation}
\mathcal H^\ell=\operatorname{Predict}(V^\ell,Q^\ell)
=\left(
\{\widehat B^\ell_{t,p},z^\ell_{t,p},q^\ell_{t,p}\}_{t,p},
\{r^\ell_t,b^\ell_t\}_{t=1}^{T}
\right).
\label{eq:gst-prediction}
\end{equation}
The fixed proposal centers and the query scale are implicit inputs. Each
$\widehat B^\ell_{t,p}$ is a complete 9-DoF cuboid. Its center is anchored
to the voted search proposal, its dimensions use the scalar query scale
as a prior, and its rotation is predicted directly in the search frame
using a continuous 6D representation~\citep{zhou6drotation}. Two
within-head residual corrections refine the cuboid before it is output (Appendix~\ref{app:denoising_details}).
 
Association logit $z^\ell_{t,p}$ measures the proposal's match to the query,
whereas quality $q^\ell_{t,p}\in(0,1)$ estimates geometric agreement with
the target cuboid. Frame-level response and absence logits, $r^\ell_t$
and $b^\ell_t$, provide positive and negative evidence of target presence.
The initial output $\mathcal H^0$ supplies the first QTM round. Prediction,
within-head refinement, and quality targets are detailed in
Appendices~\ref{app:feature_details}--\ref{app:temporal_reasoning_details}.
 
\subsection{Query--Tube--Memory Refinement}
\label{sec:gst_qtm}
 
QTM selects reference observations using both proposal-level and
frame-level evidence, then uses their predicted geometry to refine the query
and proposal features.
 
\paragraph{Reference selection.}
We combine proposal association and quality with frame response and absence:
\begin{equation}
\begin{aligned}
\rho_t^\ell
&=\tfrac12\max_p\left[z^\ell_{t,p}
+\lambda_q\log(q^\ell_{t,p}+\epsilon)\right]
+\tfrac12\left[r^\ell_t
-\lambda_b\operatorname{softplus}(b^\ell_t)\right],\\
\mathcal S^\ell
&=\operatorname{Select}(\mathcal H^\ell,\boldsymbol\rho^\ell).
\end{aligned}
\label{eq:gst-reference}
\end{equation}
Here, $\lambda_q$ and $\lambda_b$ weight the quality and absence terms,
$\epsilon$ is a small constant, and $\sigma$ denotes the sigmoid; the
weights are listed in Appendix~\ref{app:implementation}.
The maximizing proposal supplies the reference cuboid in each frame.
Among the three highest-scoring frames, we retain those satisfying
$\sigma(\rho_t^\ell)\geq0.7$, always keeping the strongest frame as a
fallback. References may come from any occurrence; the latest occurrence
is selected only during final decoding.
 
\paragraph{Box-local memory.}
For selected cuboid
$B_j^\ell=(\mathbf c_j^\ell,\mathbf s_j^\ell,\mathbf R_j^\ell)$ in reference
frame $t_j^\ell$, we pool proposal evidence using cuboid-normalized distances:
\begin{equation}
\mathbf m_j^\ell
=\sum_{p=1}^{P}
\operatorname{softmax}_{p}\!\left(
-\left\|
\operatorname{diag}(\mathbf s_j^\ell/2)^{-1}
(\mathbf R_j^\ell)^\top
(\mathbf p_{t_j^\ell,p}-\mathbf c_j^\ell)
\right\|_2^2
\right)\mathbf u^\ell_{t_j^\ell,p}.
\label{eq:gst-memory}
\end{equation}
Here, $\mathbf u^\ell_{t,p}$ is the query-conditioned proposal token, the
fusion of the corresponding proposal feature in $V^\ell$ with the mean of
the query tokens $Q^\ell$, from which the shared head predicts
(Appendix~\ref{app:feature_details}). The 
center, size, and orientation of the cuboid jointly determine the pooling weights. 
The softmax runs over all proposals in the reference frame, so every proposal
contributes with a weight set by its position relative to the cuboid. The pooled tokens, augmented with selected geometry,
normalized frame position, and reference confidence, form $\mathcal M^\ell$.
 
\paragraph{Feature refinement.}
The Query Updater applies self-attention and memory cross-attention, with
a gated residual anchored to $Q^0$ to retain the original target
information. The Tube Updater writes the same memory into proposal
features, using box-local compatibility, temporal distance, reference
confidence, and validity to gate the updates:
\begin{equation}
\begin{aligned}
Q^{\ell+1}
&=\operatorname{QueryUpdater}(Q^\ell,\mathcal M^\ell,Q^0),\\
V^{\ell+1}
&=\operatorname{TubeUpdater}(V^\ell,\mathcal M^\ell),\\
\mathcal H^{\ell+1}
&=\operatorname{Predict}(V^{\ell+1},Q^{\ell+1}).
\end{aligned}
\label{eq:gst-update}
\end{equation}
Proposal centers and the query scale stay fixed across rounds, so each round
rewrites only $Q^\ell$ and $V^\ell$. We apply two rounds. The second round
rebuilds the references and the memory from $\mathcal H^1$ and produces
$\mathcal H^2$. Further details are given in
Appendix~\ref{app:temporal_reasoning_details}.
 
\subsection{Response-Tube Decoding}
\label{sec:gst_decoding}
 
We decode a cuboid path and a response interval from $\mathcal H^2$:
\begin{equation}
\widehat{\mathcal T}
=\operatorname{TubeDecode}(\mathcal H^2)
=\{(t,\widehat B_t):t\in\widehat{\mathcal R}\}.
\label{eq:gst-decoding}
\end{equation}
The path decoder retains the eight highest-association cuboids per frame.
Dynamic programming selects one per frame using association-derived unary
costs and low-weight center, log-size, and rotation continuity costs.
The continuity costs are computed in the camera coordinates of each frame. Quality guides QTM reference selection but does not enter
the final path score. For the interval, we median-filter
$\sigma(r_t^2-\lambda_b\operatorname{softplus}(b_t^2))$, select the latest
peak satisfying the relative peak threshold, and expand it in both
directions to obtain $\widehat{\mathcal R}$. The costs and thresholds are
specified in Appendix~\ref{app:inference_details}.
 
\subsection{Optimization with ST-D9O}
\label{sec:gst_training}
 
Intermediate cuboids determine how QTM pools features. We therefore
supervise their complete geometry together with the individual parameters. The training-only Spatio-Temporal Differentiable 9-DoF
Operator (ST-D9O) maps a cuboid to its corners, sampled surfaces,
signed-distance field, and soft occupancy:
$\Gamma(B)=(\mathcal C(B),\mathcal S(B),\operatorname{SDF}_B,\operatorname{Occ}_B)$,
where $\operatorname{Occ}_B(\mathbf x)
=\sigma[-\operatorname{SDF}_B(\mathbf x)/\tau_{\mathrm{occ}}]$ 
and $\tau_{\mathrm{occ}}$ is the occupancy temperature.
Grouping discrepancies by geometric role gives
\begin{equation}
D_{\mathrm{9D}}(\widehat B,B)
=D_{\mathrm{param}}(\widehat B,B)
+D_{\mathrm{boundary}}(\widehat B,B)
+D_{\mathrm{SDF}}(\widehat B,B)
+D_{\mathrm{overlap}}(\widehat B,B).
\label{eq:gst-discrepancy}
\end{equation}
Each group includes its configured weights. The parameter group compares
normalized center, log-size, log-volume, and rotation; the boundary group
compares corners and surface samples. The SDF group measures bidirectional
surface-to-field discrepancies, and the overlap group approximates cuboid
overlap using soft occupancy. Rotation and boundary comparisons account
for cuboid-equivalent local-axis flips. The operator is differentiable
almost everywhere, and its soft overlap approximates 9-DoF IoU through sampled
occupancy. Appendix~\ref{app:std9o_details} gives the full construction.
 
At every stage, $D_{\mathrm{9D}}$ supervises the selected prediction in each
positive frame and nearby positive proposals. With low-weight temporal
regularization over contiguous positive runs, the geometry objective is
\begin{equation}
\mathcal L_{\mathrm{geo}}^\ell
=\mathcal L_{\mathrm{selected}}^\ell
+0.25\mathcal L_{\mathrm{dense}}^\ell
+0.05\mathcal L_{\mathrm{temp}}^\ell.
\label{eq:gst-geometry-objective}
\end{equation}
The complete training objective is
\begin{equation}
\mathcal L=\mathcal L_{\mathrm{init}}
+\sum_{\ell=0}^{2}w_\ell
\left(
\mathcal L_{\mathrm{disc}}^\ell
+0.5\mathcal L_{\mathrm{quality}}^\ell
+\mathcal L_{\mathrm{geo}}^\ell
\right),
\qquad
(w_0,w_1,w_2)=(3,2,1)/6.
\label{eq:gst-objective}
\end{equation}
Here, $\mathcal L_{\mathrm{init}}$ supervises point objectness and voting,
and $\mathcal L_{\mathrm{disc}}^\ell$ groups the weighted proposal
classification, ranking, response, and absence objectives.
$\mathcal L_{\mathrm{quality}}^\ell$ uses detached geometric targets for
positives and zero targets for valid negatives. Appendix~\ref{app:training_details} expands these groups.
GT-near proposals provide additional positive training samples and are
progressively removed (Appendix~\ref{app:gt_near_warmup}). Neither these
auxiliary proposals nor ST-D9O is used at inference.

\section{Experiments}
\label{sec:experiments}
 
\subsection{Experimental Setup}
\label{sec:experimental_setup}
 
\paragraph{Dataset and metrics.}
We evaluate on the RGB--point-cloud setting of 3DVQL~\citep{laf}, which
contains 2,002 multimodal sequences from 38 object categories across 18
environments. The official split contains 1,601 sequences, 131.4K frames,
and 5,157 response tracks for training, and 401 sequences, 39.6K frames,
and 1,319 response tracks for testing. The task retrieves the latest
contiguous target occurrence and its frame-wise 9-DoF cuboids. We report
temporal average precision (tAP), mean spatio-temporal average precision
(stAP), Recovery, and Success, together with threshold-specific
$\mathrm{tAP}_{0.25}$, $\mathrm{stAP}_{0.05}$, and $\mathrm{stAP}_{0.25}$.
Among these, $\mathrm{stAP}_{0.25}$ and Recovery (frame IoU $\geq0.5$) are 
the strictest measures of cuboid accuracy, whereas Success only requires $\mathrm{stIoU}_{3D}\geq0.05$.
Metric definitions are given in
Appendix~\ref{app:evaluation_protocol}.
 
\paragraph{Implementation.}
We sample $T=30$ search frames and 3,072 points per frame. A frozen DINOv2
ViT-B/14~\citep{dinov2} extracts RGB features, and a lightweight cascaded
EdgeConv backbone~\citep{dgcnn} is trained from scratch for point-cloud
features. The encoder retains 96 query candidates and 192 search
candidates per frame. The query is summarized into 32 tokens, and
proposal initialization generates 96 proposals per frame. All trainable
features have dimension 96. The base configuration uses 400 epochs,
four RTX~4090 GPUs, an effective batch size of 8, and AdamW with learning
rate $10^{-4}$ and weight decay $5\times10^{-3}$. Detailed settings and
baseline adaptations are given in Appendix~\ref{app:implementation}.
 
\paragraph{Reporting protocol.}
The complete-model results in Table~\ref{tab:gst-main} report the mean and
standard deviation over seeds 42, 43, and 44; the seed-level results are
listed in Table~\ref{tab:app_seed_results}. All component and stage-wise
analyses branch from the seed-42 run, which serves as their shared control
(0.777 tAP, 0.266 stAP). Each ablation is a single run. We therefore draw
conclusions from differences that clearly exceed the seed-level spread of the
complete model (0.004 stAP and 0.013 tAP), and we describe smaller
differences as consistent in direction.
 
\subsection{Comparison with Existing Methods}
\label{sec:main_comparison}
 
We compare with AF, GAF, PAF, and LaF as reported by the 3DVQL
benchmark~\citep{laf}. PRVQL-3D is our adaptation of PRVQL~\citep{prvql}
to RGB--point-cloud input and frame-wise 9-DoF prediction; it shares the
input, split, and evaluator with PGL-3D and differs in architecture
(Appendix~\ref{app:baseline_reproduction}). Table~\ref{tab:gst-main} reports
system-level performance. Compared with LaF, PGL-3D increases
$\mathrm{stAP}_{0.05}$ from 0.222 to 0.746 and Success from 46.0\% to
78.8\%. Compared with PRVQL-3D, $\mathrm{stAP}_{0.25}$ increases from
0.025 to 0.213. Across three seeds, PGL-3D obtains
$0.270\pm0.004$ mean stAP and $0.766\pm0.013$ tAP. Part of the gain over
LaF comes from the encoding and proposal pipeline: with both QTM updates
disabled (Table~\ref{tab:app_qtm_update_paths}), PGL-3D already reaches
0.178 stAP, above LaF and PRVQL-3D. Sec.~\ref{sec:ablation_study} analyzes
the contributions of QTM and ST-D9O.

\begin{table}[t]
\centering
\caption{\textbf{Comparison on 3DVQL.} AF, GAF, PAF, and LaF are
reported by \citet{laf}; $\dagger$ denotes our 3D adaptation of PRVQL~\cite{prvql}.
PGL-3D reports the three-seed mean, with standard deviations below it.
Recovery and Success are percentages; -- denotes an unreported value.}
\label{tab:gst-main}
\small
\setlength{\tabcolsep}{3pt}
\renewcommand{\arraystretch}{1.08}
\begin{tabularx}{\linewidth}{>{\raggedright\arraybackslash}Xrrrrrrr}
\toprule
\rowcolor{gray!12}
Method & tAP$\,\uparrow$ & $\mathrm{tAP}_{0.25}\,\uparrow$ &
stAP$\,\uparrow$ & $\mathrm{stAP}_{0.05}\,\uparrow$ &
$\mathrm{stAP}_{0.25}\,\uparrow$ & Rec.\ (\%)$\,\uparrow$ &
Succ.\ (\%)$\,\uparrow$ \\
\midrule
AF
& 0.181\phantom{0} & 0.442\phantom{0} & 0.003\phantom{0}
& 0.015\phantom{0} & -- & 0.093\phantom{0} & 11.693\phantom{0} \\
GAF
& 0.291\phantom{0} & 0.597\phantom{0} & 0.015\phantom{0}
& 0.075\phantom{0} & -- & 0.049\phantom{0} & 26.309\phantom{0} \\
PAF
& 0.224\phantom{0} & 0.577\phantom{0} & 0.021\phantom{0}
& 0.104\phantom{0} & -- & 0.115\phantom{0} & 32.156\phantom{0} \\
LaF
& 0.293\phantom{0} & 0.607\phantom{0} & 0.044\phantom{0}
& 0.222\phantom{0} & -- & 0.264\phantom{0} & 46.041\phantom{0} \\
\midrule
PRVQL-3D$^\dagger$
& \underline{0.546}\phantom{0} & \underline{0.685}\phantom{0}
& \underline{0.099}\phantom{0} & \underline{0.402}\phantom{0}
& \underline{0.025}\phantom{0} & \underline{2.00}\phantom{0}
& \underline{55.4}\phantom{0} \\
\midrule
\rowcolor{pglhighlight}
\textbf{PGL-3D} (mean)
& \textbf{0.766} & \textbf{0.836} & \textbf{0.270}
& \textbf{0.746} & \textbf{0.213} & \textbf{5.70}
& \textbf{78.8} \\
\rowcolor{pglhighlight}
\quad Std.
& $\pm$0.013 & $\pm$0.012 & $\pm$0.004
& $\pm$0.017 & $\pm$0.010 & $\pm$0.97 & $\pm$0.66 \\
\bottomrule
\end{tabularx}
\end{table}

\subsection{Ablation Study}
\label{sec:ablation_study}
 
\paragraph{Supervision and geometry feedback.}
Table~\ref{tab:gst-feedback} compares training-supervision and
QTM-guidance variants. Final-stage supervision only yields 0.225 stAP;
intermediate supervision without geometry feedback yields 0.246.
Geometry feedback with association-only guidance reaches 0.254, and
full PGL-3D reaches 0.266. The full model also improves
$\mathrm{stAP}_{0.25}$ from 0.188 to 0.215 over association-only guidance.
This progression is consistent with benefits from prediction-guided
feature updates and multi-signal reference selection. Complete metrics
and the definition of each variant are given in
Appendix~\ref{app:feedback_ablations}.

\begin{table}[t]
\centering
\caption{\textbf{Intermediate supervision and geometry-guided refinement.}
Each row adds one component to the row above;
Table~\ref{tab:app_variant_definitions} lists the switches of every variant.
The full-model row is the seed-42 control.}
\label{tab:gst-feedback}
\small
\setlength{\tabcolsep}{3pt}
\renewcommand{\arraystretch}{1.08}
\begin{tabularx}{\linewidth}{l*{3}{>{\raggedleft\arraybackslash}X}}
\toprule
\rowcolor{gray!12}
Variant & tAP$\,\uparrow$ & stAP$\,\uparrow$ &
$\mathrm{stAP}_{0.25}\,\uparrow$ \\
\midrule
Final-stage supervision only & 0.702 & 0.225 & 0.158 \\
Intermediate supervision & 0.721 & 0.246 & 0.179 \\
Association-only QTM
& \underline{0.745} & \underline{0.254} & \underline{0.188} \\
\rowcolor{pglhighlight}
\textbf{Full PGL-3D}
& \textbf{0.777} & \textbf{0.266} & \textbf{0.215} \\
\bottomrule
\end{tabularx}
\end{table}
 
\paragraph{Progress across prediction stages.}
Table~\ref{tab:gst-stages} evaluates the three intermediate outputs of the
complete model. From $\mathcal H^0$ to $\mathcal H^2$, stAP increases from
0.229 to 0.266 and mean frame IoU from 0.208 to 0.237. Center, size,
and rotation errors decrease from 0.190\,m, 0.245\,m, and $22.5^\circ$
to 0.158\,m, 0.209\,m, and $19.3^\circ$, respectively. Reference precision
increases from 43.8\% to 48.7\%. The first QTM round gives the larger gain
in stAP (+0.020), and the second round adds +0.017. The observed changes
involve both retrieval and cuboid geometry.
 
 
 
\begin{table}[t]
\centering
\caption{\textbf{Prediction quality across stages.} All rows are intermediate
outputs of the same trained model. Units and preferred directions appear in the headers.}
\label{tab:gst-stages}
\small
\setlength{\tabcolsep}{3pt}
\renewcommand{\arraystretch}{1.08}
\begin{tabularx}{\linewidth}{l*{7}{>{\raggedleft\arraybackslash}X}}
\toprule
\rowcolor{gray!12}
Stage & tAP$\,\uparrow$ & stAP$\,\uparrow$ &
\shortstack[r]{Frame IoU\\$\uparrow$} &
\shortstack[r]{Center err.\\(m)$\,\downarrow$} &
\shortstack[r]{Size err.\\(m)$\,\downarrow$} &
\shortstack[r]{Rot. err.\\($^\circ$)$\,\downarrow$} &
\shortstack[r]{Ref. prec.\\(\%)$\,\uparrow$} \\
\midrule
$\mathcal H^0$
& 0.716 & 0.229 & 0.208 & 0.190 & 0.245 & 22.5 & 43.8 \\
$\mathcal H^1$
& \underline{0.751} & \underline{0.249} & \underline{0.228}
& \underline{0.168} & \underline{0.218} & \underline{20.0}
& \underline{46.5} \\
\rowcolor{pglhighlight}
$\mathcal H^2$
& \textbf{0.777} & \textbf{0.266} & \textbf{0.237}
& \textbf{0.158} & \textbf{0.209} & \textbf{19.3} & \textbf{48.7} \\
\bottomrule
\end{tabularx}
\end{table}

\paragraph{Impact of ST-D9O.}
In Table~\ref{tab:gst-components}, separate center, size, and rotation
regression gives higher tAP (0.817) but lower stAP (0.183) and
$\mathrm{stAP}_{0.25}$ (0.076) than the full objective. Strong temporal
retrieval therefore does not ensure accurate cuboids. Disabling both QTM
updates lowers stAP to a similar level (0.178,
Table~\ref{tab:app_qtm_update_paths}), so accurate cuboids require both the
geometry objective and the geometry-guided feature updates. The MGIoU
variant, which replaces the soft-overlap term of ST-D9O with the MGIoU
loss, reaches 0.220 stAP and 0.160 $\mathrm{stAP}_{0.25}$, compared with
0.266 and 0.215 for full ST-D9O. The corresponding absolute gains are
0.046 and 0.055 (20.9\% and 34.4\% relative), respectively.
Removing dense supervision or temporal regularization reduces
$\mathrm{stAP}_{0.25}$ to 0.190 or 0.199. The incremental analysis in
Table~\ref{tab:app_std9o_incremental} further separates the geometry terms:
$\mathrm{stAP}_{0.25}$ rises from 0.118 with parameter terms to 0.151
with boundaries, 0.169 with SDF, and 0.196 with soft overlap, before
dense and temporal supervision bring it to 0.215.

\paragraph{Other components.}
Removing RGB raises tAP to 0.803 and lowers $\mathrm{stAP}_{0.25}$ to 0.181;
removing Temporal Feature Aggregation lowers both metrics. Removing cuboid
quality gives the highest tAP in Table~\ref{tab:gst-components}, yet reduces
$\mathrm{stAP}_{0.25}$ to 0.154. Target Absence and Temporal Feature
Refinement also benefit strict localization. Two geometric components,
cuboid quality and ST-D9O, thus raise strict localization while lowering
tAP. A plausible explanation is that the shared head serves both retrieval
and geometry: cuboid quality steers reference selection toward proposals with
reliable geometry, and ST-D9O pulls the shared features toward geometric
agreement, so both sharpen size and orientation and both can cost some
association and response discrimination. We report this trade-off as an
observation, since isolating its cause would require repeated runs of each
variant. Separate Query- and Tube-Updater variants reach
0.232 and 0.221 stAP, compared with 0.266 for both updates
(Table~\ref{tab:app_qtm_update_paths}). Removing GT-near warm-up proposals
reduces $\mathrm{stAP}_{0.25}$ to 0.187. Complete component results,
temporal-descriptor diagnostics, and decoding ablations are reported in
Appendix~\ref{app:additional_experiments}.
 
 
\begin{table}[t]
\centering
\caption{\textbf{Component ablations and geometry-objective comparison.}
The full-model row is the seed-42 control. Complete results appear in Table~\ref{tab:app_complete_ablations}.}
\label{tab:gst-components}
\small
\setlength{\tabcolsep}{3pt}
\renewcommand{\arraystretch}{1.08}
\begin{tabularx}{\linewidth}{l*{4}{>{\raggedleft\arraybackslash}X}}
\toprule
\rowcolor{gray!12}
Variant & tAP$\,\uparrow$ & stAP$\,\uparrow$ &
$\mathrm{stAP}_{0.05}\,\uparrow$ & $\mathrm{stAP}_{0.25}\,\uparrow$ \\
\midrule
\rowcolor{pglhighlight}
\textbf{Full PGL-3D}
& 0.777 & \textbf{0.266} & \underline{0.729} & \textbf{0.215} \\
\midrule
\multicolumn{5}{l}{\textit{Multimodal encoding}} \\
\hspace*{0.5em}w/o RGB
& 0.803 & 0.235 & 0.689 & 0.181 \\
\hspace*{0.5em}w/o Temporal Feature Aggregation
& 0.735 & 0.232 & 0.669 & 0.176 \\
\addlinespace[2pt]
\multicolumn{5}{l}{\textit{QTM}} \\
\hspace*{0.5em}w/o cuboid quality
& \textbf{0.831} & 0.246 & \textbf{0.735} & 0.154 \\
\hspace*{0.5em}w/o Target Absence
& 0.741 & 0.254 & 0.685 & \underline{0.202} \\
\hspace*{0.5em}w/o Temporal Feature Refinement
& 0.728 & 0.227 & 0.666 & 0.174 \\
\addlinespace[2pt]
\multicolumn{5}{l}{\textit{Geometry objective}} \\
\hspace*{0.5em}Separate Regression
& \underline{0.817} & 0.183 & 0.672 & 0.076 \\
\hspace*{0.5em}w/ MGIoU
& 0.760 & 0.220 & 0.680 & 0.160 \\
\hspace*{0.5em}w/o Dense Loss
& 0.763 & 0.251 & 0.706 & 0.190 \\
\hspace*{0.5em}w/o Temporal Loss
& 0.770 & \underline{0.258} & 0.712 & 0.199 \\
\bottomrule
\end{tabularx}
\end{table}
 
\subsection{Transfer of ST-D9O to 3D Tracking}
\label{sec:gst_transfer}
 
We evaluate ST-D9O with PROT3D on the GSOT3D point-cloud
benchmark~\citep{gsot3d}. Starting from our reproduction, we replace only
the box-regression objective and retain the architecture, data sampling,
training schedule, and inference procedure. ST-D9O increases mAO from
21.63\% to 25.78\%, $\mathrm{mSR}_{50}$ from 19.38\% to 23.61\%, and
$\mathrm{mSR}_{75}$ from 5.08\% to 6.23\%
(Table~\ref{tab:gst-transfer}). This supports transfer to another
architecture in the tested tracking setting. The complete comparison
with published GSOT3D baselines is reported in
Table~\ref{tab:app_gsot_full}.
 
 
\begin{table}[t]
\centering
\caption{\textbf{Replacing the geometry objective of PROT3D on GSOT3D.}
The reported result is from \citet{gsot3d}; the last two rows form the
controlled comparison. All metrics are percentages.}
\label{tab:gst-transfer}
\small
\setlength{\tabcolsep}{3pt}
\renewcommand{\arraystretch}{1.08}
\begin{tabularx}{\linewidth}{l*{3}{>{\raggedleft\arraybackslash}X}}
\toprule
\rowcolor{gray!12}
Method & mAO$\,\uparrow$ & $\mathrm{mSR}_{50}\,\uparrow$ &
$\mathrm{mSR}_{75}\,\uparrow$ \\
\midrule
PROT3D (reported)
& \underline{21.97} & \underline{19.76} & \underline{5.22} \\
\midrule
PROT3D (our reproduction)
& 21.63 & 19.38 & 5.08 \\
\rowcolor{pglhighlight}
\textbf{PROT3D + ST-D9O}
& \textbf{25.78} & \textbf{23.61} & \textbf{6.23} \\
\bottomrule
\end{tabularx}
\end{table}
 
\subsection{Efficiency and Qualitative Results}
\label{sec:efficiency_qualitative}

\begin{figure}[t]
\centering
\includegraphics[width=\linewidth]{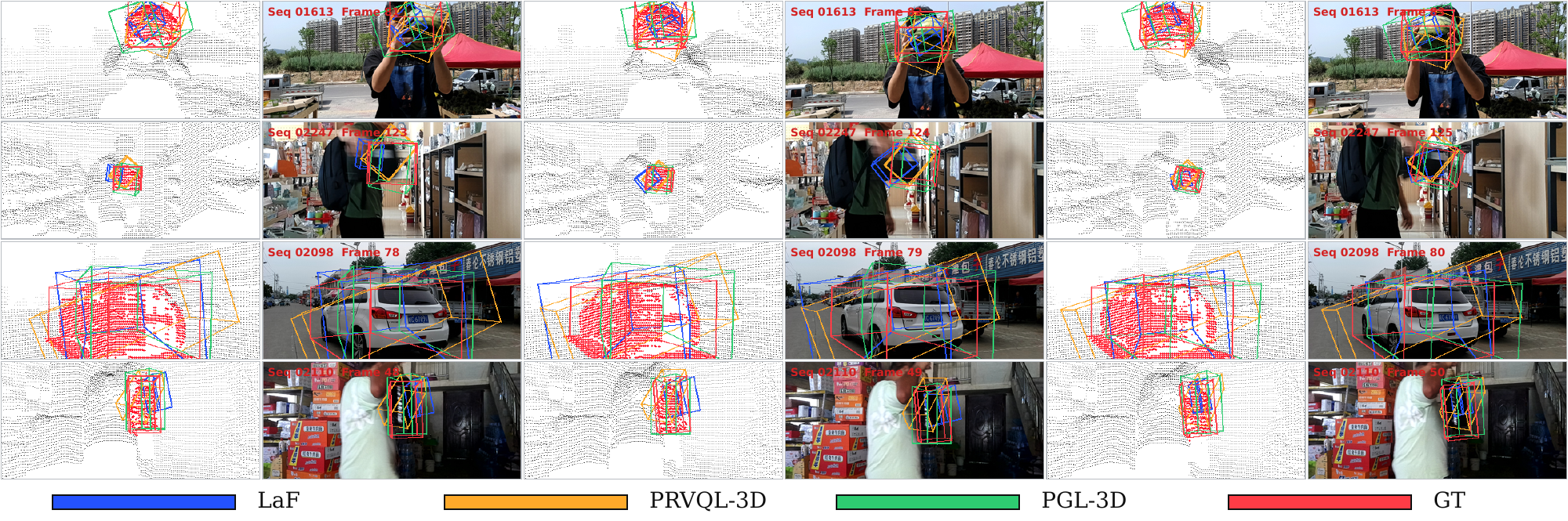}
\caption{Qualitative comparison of frame-wise 9-DoF localization.
Each row shows three frames. Blue, orange, green, and red boxes denote
LaF, PRVQL-3D, PGL-3D, and ground truth, respectively.}
\label{fig:response_tube_analysis}
\end{figure}
 
\paragraph{Efficiency.}
On one RTX~4090 with batch size 1, PGL-3D processes 30 frames in
1.99\,s (15.1 FPS) with 9.8\,GB peak inference memory. Measurements
use cached DINOv2 features and exclude disk I/O. ST-D9O is disabled at inference.
Table~\ref{tab:app_efficiency} lists LaF and PRVQL-3D under the same
protocol; the three methods process different numbers of frames.
 
\paragraph{Qualitative results.}
Figure~\ref{fig:response_tube_analysis} compares frame-wise localization
with LaF and PRVQL-3D on four sequences. The examples illustrate differences
in cuboid position, extent, and orientation that temporal metrics alone do
not capture. The LaF and PRVQL-3D cuboids often cover the target region but
deviate from the ground truth in extent or orientation, whereas the PGL-3D
cuboid stays close to the ground truth across the three consecutive frames.
This behavior matches the stage-wise decrease of size and rotation error in
Table~\ref{tab:gst-stages} and the gap in $\mathrm{stAP}_{0.25}$ between the
methods. Stage-wise visualizations and failure cases are provided in
Appendix~\ref{app:qualitative}; the remaining failures involve severe
occlusion, extreme point sparsity, and geometrically similar distractors.

\section{Conclusion}
\label{sec:conclusion}
This paper presents PGL-3D for 3D Visual Query Localization. Intermediate
9-DoF cuboids, selected by the four reliability signals of QTM, define soft
pooling into Box-Local Memory and refine the query and proposal features
before final decoding, while ST-D9O supervises the complete cuboid geometry
at every stage without inference-time cost. On 3DVQL, PGL-3D raises mean
stAP from 0.044 (LaF) to 0.270; geometry-guided updates improve over
intermediate supervision alone (0.246 to 0.266), and removing either QTM or
ST-D9O lowers stAP to about 0.18. ST-D9O also improves PROT3D on GSOT3D
(21.63\% to 25.78\% mAO). Intermediate geometry thus serves not only as a
prediction target but as a condition for subsequent feature aggregation and
refinement.
 
%
 
 

\newpage
\section*{Reproducibility Statement}
Implementation details, training configurations, and evaluation protocols are
provided in the main paper and appendix. We will release the training and
evaluation code upon publication.
 
\section*{AI Use Statement}
Generative AI tools were used to assist with language editing, LaTeX formatting,
literature organization, and figure prototyping. All AI-assisted content was
reviewed by the authors. The authors take full responsibility for the technical
claims, experiments, results, and final manuscript.

\bibliography{refs}
\bibliographystyle{iclr2027_conference}
 
\appendix

\section{Additional Method Details}
\label{app:method_details}
 
This supplement follows the execution order and notation in
Sec.~\ref{sec:method}. It gives the implementation details omitted from the
main paper. These include feature extraction and fusion, Temporal Feature
Aggregation, 3D Feature Learning, QTM, ST-D9O, training, and Response-Tube
Decoding. It also reports complete ablation metrics and additional analyses.
 
\subsection{Multimodal Encoding}
\label{app:evidence_details}
 
\paragraph{Coordinate convention.}
The independent query is represented in $\mathcal F^q$. Search frame $t$
uses the calibrated camera coordinate system $\mathcal F_t^s$, where its
9-DoF annotation and prediction are defined. The calibration maps each
retained 3D point to the RGB camera for feature sampling. No query-to-search
or inter-frame transform is estimated.
 
\paragraph{Query canonicalization.}
The query is
$\mathcal Q=(I^q,P^q,\mathcal C^q,B^q)$, where
$B^q=(\mathbf c^q,\mathbf s^q,\mathbf R^q)$ is its annotated 9-DoF cuboid.
Because the query and search observations are collected independently, the
query center and orientation are used only to express query points in the
local coordinate system of the query cuboid. For
$\mathbf x_i^q\in P^q$, we define
\begin{equation}
\mathbf u_i^q
=
\operatorname{diag}(\mathbf s^q/2)^{-1}
(\mathbf R^q)^\top
(\mathbf x_i^q-\mathbf c^q).
\label{eq:app_query_canonicalization}
\end{equation}
The query geometric stem receives
$[\mathbf u_i^q,\|\mathbf u_i^q\|_2]$, whereas the search geometric stem
receives normalized coordinates in $\mathcal F_t^s$ and local relative
offsets. In particular, no search feature contains
$\mathbf x_{t,i}-\mathbf c^q$, and neither $\mathbf c^q$ nor
$\mathbf R^q$ is passed to a search-side coordinate operation.
 
The two branch-specific stems map their inputs to a common feature dimension,
after which a shared lightweight cascaded EdgeConv-style local graph encoder
performs dynamic KNN aggregation and progressive point sampling. The encoder
retains 96 query candidates and 192 search candidates per frame.
 
\paragraph{RGB point painting.}
The query and search branches share a frozen DINOv2 image backbone but use
branch-specific RGB projection and gating layers. We write the search branch
explicitly. Let
$\mathcal C_t=(\mathbf K_t,\mathbf T_t^{c_t\leftarrow s_t})$, where
$\mathbf K_t$ is the intrinsic matrix and
$\mathbf T_t^{c_t\leftarrow s_t}$ maps the calibrated search coordinates in
$\mathcal F_t^s$ to the RGB-camera frame. A search point is projected by
\begin{equation}
\begin{bmatrix}
\widetilde{\mathbf x}_{t,i}^{cam}\\1
\end{bmatrix}
=
\mathbf T_t^{c_t\leftarrow s_t}
\begin{bmatrix}
\mathbf x_{t,i}\\1
\end{bmatrix},
\qquad
\mathbf u_{t,i}^{img}
=
\pi\!\left(
\mathbf K_t\widetilde{\mathbf x}_{t,i}^{cam}
\right),
\label{eq:app_rgb_projection}
\end{equation}
where $\widetilde{\mathbf x}_{t,i}^{cam}\in\mathbb R^3$ is the Cartesian
RGB-camera coordinate and $\pi(\cdot)$ denotes perspective division.
The RGB descriptor
$\mathbf a_{t,i}$ is bilinearly sampled from the frozen image feature map.
 
Let $\mathbf f_{t,i}^{pc}$ denote the point feature. RGB and point features
are fused through
\begin{equation}
\begin{aligned}
\widetilde{\mathbf a}_{t,i}
&=
\mathbf W_{\mathrm{rgb}}^{s}\mathbf a_{t,i},\\
\boldsymbol\gamma_{t,i}
&=
\sigma\!\left(
\phi_g^{s}
[
\mathbf f_{t,i}^{pc},
\widetilde{\mathbf a}_{t,i},
\mathbf f_{t,i}^{pc}\odot\widetilde{\mathbf a}_{t,i}
]
\right),\\
\mathbf e_{t,i}
&=
\mathbf f_{t,i}^{pc}
+
\lambda_{\mathrm{rgb}}
\boldsymbol\gamma_{t,i}\odot\widetilde{\mathbf a}_{t,i}.
\end{aligned}
\label{eq:app_rgb_fusion}
\end{equation}
The query branch uses its original calibrated coordinates for RGB sampling
and pairs the sampled RGB descriptors with the canonical geometric features
from Eq.~\ref{eq:app_query_canonicalization}. It uses separate parameters
$\mathbf W_{\mathrm{rgb}}^{q}$ and $\phi_g^{q}$.
 
Query candidates satisfying
$\|\mathbf u_i^q\|_\infty\leq1.5$ are summarized into
$K_q=32$ tokens,
$Q^0=\{\mathbf q_k^0\}_{k=1}^{K_q}$, with pooled representation
$\bar{\mathbf q}^{\,0}
=K_q^{-1}\sum_{k=1}^{K_q}\mathbf q_k^0$.
The search branch retains 192 point-grounded multimodal candidates per frame.
 
\subsection{Temporal Feature Aggregation}
\label{app:temporal_point_memory}
 
Temporal context is incorporated before proposal generation. Let
$\bar{\mathbf x}_{t,i}$ denote the point coordinate normalized by the
benchmark sampling range, and let
$\Omega_t=\{\tau:0<|t-\tau|\leq R\}$ be the neighboring-frame set. For each
neighboring frame, cross-frame KNN is computed in the normalized point
coordinates,
\begin{equation}
\mathcal N_{t,i}^{\tau}
=
\operatorname{KNN}_{j}^{K}
\left(
\|\bar{\mathbf x}_{\tau,j}-\bar{\mathbf x}_{t,i}\|_2
\right),
\qquad R=5,\quad K=8.
\label{eq:app_spacetime_neighbors}
\end{equation}
For $j\in\mathcal N_{t,i}^{\tau}$, define the relative coordinate and signed
time offset as
$\Delta\bar{\mathbf x}_{t,i,\tau,j}
=\bar{\mathbf x}_{\tau,j}-\bar{\mathbf x}_{t,i}$ and
$\Delta\tau=(\tau-t)/\max(T-1,1)$. The temporal edge descriptor is
\begin{equation}
\mathbf d_{t,i,\tau,j}
=
\left[
\mathbf e_{t,i},
\mathbf e_{\tau,j}-\mathbf e_{t,i},
\Delta\bar{\mathbf x}_{t,i,\tau,j},
\|\Delta\bar{\mathbf x}_{t,i,\tau,j}\|_2,
\Delta\tau
\right].
\label{eq:app_temporal_edge}
\end{equation}
Messages are pooled within each source frame and averaged across the temporal
window,
\begin{equation}
\mathbf r_{t,i}^{st}
=
\frac{1}{|\Omega_t|}
\sum_{\tau\in\Omega_t}
\operatorname*{Pool}_{j\in\mathcal N_{t,i}^{\tau}}
\phi_{\mathrm{edge}}(\mathbf d_{t,i,\tau,j}).
\label{eq:app_temporal_aggregate}
\end{equation}
When $t>1$, the nearest point in the previous frame provides an additional
cross-frame displacement descriptor. With
$j_{t,i}^{\mathrm{prev}}=
\arg\min_j\|\bar{\mathbf x}_{t-1,j}-\bar{\mathbf x}_{t,i}\|_2$, we use
\begin{equation}
\mathbf m_{t,i}^{\mathrm{prev}}
=
\left[
\bar{\mathbf x}_{t-1,j_{t,i}^{\mathrm{prev}}}
-
\bar{\mathbf x}_{t,i},
\left\|
\bar{\mathbf x}_{t-1,j_{t,i}^{\mathrm{prev}}}
-
\bar{\mathbf x}_{t,i}
\right\|_2
\right],
\label{eq:app_previous_displacement}
\end{equation}
and set it to zero for the first frame. A frame-local descriptor combines the
normalized current coordinate, a sinusoidal time basis $\mathbf b_t$, and the
previous-frame displacement:
\begin{equation}
\mathbf s_{t,i}^{st}
=
\phi_{\mathrm{st}}
\left[
\bar{\mathbf x}_{t,i},
\mathbf b_t,
\mathbf m_{t,i}^{\mathrm{prev}}
\right].
\label{eq:app_local_spacetime_descriptor}
\end{equation}
The local and cross-frame messages are merged through a gated residual update,
\begin{equation}
\begin{aligned}
\mathbf h_{t,i}^{st}
&=\mathbf s_{t,i}^{st}+\mathbf r_{t,i}^{st},\\
\boldsymbol\eta_{t,i}^{st}
&=\sigma\!\left(
\phi_{\mathrm{stg}}
[\mathbf e_{t,i},\mathbf h_{t,i}^{st},
\mathbf e_{t,i}\odot\mathbf h_{t,i}^{st}]
\right),\\
\widetilde{\mathbf e}_{t,i}
&=\operatorname{LN}\!\left(
\mathbf e_{t,i}
+\lambda_{\mathrm{st}}
\boldsymbol\eta_{t,i}^{st}\odot\mathbf h_{t,i}^{st}
\right).
\end{aligned}
\label{eq:app_temporal_update}
\end{equation}
The relative-coordinate descriptors provide short-range cues in the
observation space of each frame. The enriched
evidence is
$E_t=\{(\mathbf x_{t,i},\widetilde{\mathbf e}_{t,i})\}_{i=1}^{M}$ and
$E_{1:T}=\{E_t\}_{t=1}^{T}$.
 
\subsection{3D Feature Learning}
\label{app:feature_details}
 
This section details how enriched point evidence is organized into
proposal-level 3D features and how cuboids and scores are predicted from those
features. The features $V^\ell$ are updated by QTM, whereas
$\mathcal H^\ell$ denotes the corresponding cuboid and score predictions.
 
\paragraph{Query-conditioned voting and proposal sampling.}
For each enriched search point, the query-conditioned seed feature is
\begin{equation}
\mathbf z_{t,i}^{pt}
=
\phi_{\mathrm{seed}}
[
\widetilde{\mathbf e}_{t,i},
\bar{\mathbf q}^{\,0},
\widetilde{\mathbf e}_{t,i}\odot\bar{\mathbf q}^{\,0}
].
\label{eq:app_seed_feature}
\end{equation}
The voting head predicts a validity logit and a bounded center offset in
$\mathcal F_t^s$:
\begin{equation}
o_{t,i}^{pt}
=
\mathbf W_o\mathbf z_{t,i}^{pt},
\qquad
\mathbf y_{t,i}
=
\mathbf x_{t,i}
+
\alpha_p\tanh(
\mathbf W_p\mathbf z_{t,i}^{pt}
).
\label{eq:app_point_vote}
\end{equation}
The vote target indicates whether $\mathbf y_{t,i}$ lies in the configured
positive region around the ground-truth center. Of the $P$ proposal centers,
half are selected from the highest-scoring votes, while the remaining half
are obtained by applying farthest-point sampling to the voted centers.
 
\paragraph{Local evidence aggregation.}
Let
$\bar s^q=(s_x^qs_y^qs_z^q)^{1/3}$ denote the rigid-invariant query scale.
For proposal center $\mathbf p_{t,p}$ and scale multiplier $\gamma_m$, the
within-frame neighborhood selects up to 32 nearest candidates within the
query-scaled radius:
\begin{equation}
\mathcal N_{t,p}^{(m)}
=
\operatorname{TopK}_{i:\,
\|\mathbf x_{t,i}-\mathbf p_{t,p}\|_2
\leq\gamma_m\bar s^q}
\|\mathbf x_{t,i}-\mathbf p_{t,p}\|_2.
\label{eq:app_search_neighborhood}
\end{equation}
Here, $\operatorname{TopK}$ selects the smallest distances. All neighborhoods
contain only points observed in the current frame and are defined in
$\mathcal F_t^s$; their orientation is independent of $\mathbf R^q$.
Local graph aggregation, max pooling, and query conditioning produce the
proposal-level 3D feature $\mathbf v_{t,p}^{0}$. We use
$V^\ell=\{\mathbf v_{t,p}^{\ell}\}_{t,p}$ for the proposal features at
prediction stage $\ell$.
 
At prediction stage $\ell$, let
$\bar{\mathbf q}^{\,\ell}=K_q^{-1}\sum_{k=1}^{K_q}\mathbf q_k^\ell$ be the
mean of the current query tokens. The proposal feature and pooled query
representation are fused as
\begin{equation}
\mathbf u_{t,p}^{\ell}
=
\phi_{\mathrm{fuse}}
[
\mathbf v_{t,p}^{\ell},
\bar{\mathbf q}^{\,\ell},
\mathbf v_{t,p}^{\ell}\odot\bar{\mathbf q}^{\,\ell},
\log\bar s^q
].
\label{eq:app_prediction_token}
\end{equation}
The prediction interface receives no raw query center or orientation.
 
\paragraph{9-DoF prediction and refinement.}
The shared heads predict association logit $z_{t,p}^{\ell}$, initial quality
logit $\kappa_{t,p}^{\ell,0}$, and complete cuboid parameters:
\begin{equation}
\begin{aligned}
\mathbf c_{t,p}^{\ell,0}
&=
\mathbf p_{t,p}
+
\alpha_c\tanh(
\mathbf W_c\mathbf u_{t,p}^{\ell}
),\\
\mathbf s_{t,p}^{\ell,0}
&=
\bar s^q
\exp\!\left[
\alpha_s\tanh(
\mathbf W_s\mathbf u_{t,p}^{\ell}
)
\right],\\
\mathbf R_{t,p}^{\ell,0}
&=
\operatorname{Rot6DToSO3}
(
\mathbf W_R\mathbf u_{t,p}^{\ell}
).
\end{aligned}
\label{eq:app_search_frame_decode}
\end{equation}
The scalar $\bar s^q$ is broadcast to three dimensions. The center is anchored
to a voted search proposal, anisotropic size is predicted around the
query-scale initialization, and orientation is predicted directly in the
search coordinate frame.
 
For
$\mathbf W_R\mathbf u_{t,p}^{\ell}
=[\mathbf a;\mathbf b]\in\mathbb R^6$, the standard continuous 6D conversion
uses
\begin{equation}
\begin{aligned}
\mathbf r_1
&=
\frac{\mathbf a}{\|\mathbf a\|_2},\\
\mathbf r_2
&=
\frac{
\mathbf b-(\mathbf r_1^\top\mathbf b)\mathbf r_1
}{
\|
\mathbf b-(\mathbf r_1^\top\mathbf b)\mathbf r_1
\|_2
},\\
\mathbf r_3
&=
\mathbf r_1\times\mathbf r_2,
\qquad
\operatorname{Rot6DToSO3}
(
[\mathbf a;\mathbf b]
)
=
[
\mathbf r_1,\mathbf r_2,\mathbf r_3
].
\end{aligned}
\label{eq:app_rot6d}
\end{equation}
 
\paragraph{Response and absence scores.}
The frame representation pools proposal features and the current query:
\begin{equation}
\mathbf f_t^\ell
=
[
\max_p\mathbf u_{t,p}^{\ell},
\operatorname{mean}_p\mathbf u_{t,p}^{\ell},
\bar{\mathbf q}^{\,\ell}
].
\label{eq:app_frame_feature}
\end{equation}
A local head and temporal 1D convolution predict the base response logit
$r_t^\ell$. An absence head cross-attends the frame representation to learned
absence tokens and predicts $b_t^\ell$. The final presence logit used for
response decoding is
\begin{equation}
\nu_t^\ell
=
r_t^\ell
-
\lambda_b
\operatorname{softplus}
(
b_t^\ell
).
\label{eq:app_presence_logit}
\end{equation}
The absence objective combines binary classification with a bidirectional
presence--absence margin. The refined cuboids, proposal-level scores, and frame-level response and
absence logits together define
\[
\mathcal H^\ell
=
\left(
\{\widehat B^\ell_{t,p},z^\ell_{t,p},q^\ell_{t,p}\}_{t,p},
\{r^\ell_t,b^\ell_t\}_{t}
\right).
\]
 
\subsection{3D Box Refinement}
\label{app:denoising_details}
 
The refiner performs $K_{\mathrm{den}}=2$ deterministic correction steps and
is trained in a denoising manner (hence the subscript): after the configured
activation iteration, bounded perturbations are added to the input cuboids
and increased through a linear warm-up, so the correction steps learn to
recover from perturbed inputs.
 
At step $k$, the current cuboid, proposal token, pooled query, association
logit, presence logit, and step embedding produce a correction token
$\mathbf d_{t,p}^{\ell,k}$. With
$\beta_k=(K_{\mathrm{den}}-k)/K_{\mathrm{den}}$ for $k=0,1$, the update is
\begin{equation}
\begin{aligned}
\mathbf c_{t,p}^{\ell,k+1}
&=
\mathbf c_{t,p}^{\ell,k}
+
\beta_k\alpha_c^{\mathrm{den}}
\tanh(
\mathbf W_c^{\mathrm{den}}
\mathbf d_{t,p}^{\ell,k}
),\\
\mathbf s_{t,p}^{\ell,k+1}
&=
\mathbf s_{t,p}^{\ell,k}
\odot
\exp\!\left[
\beta_k\alpha_s^{\mathrm{den}}
\tanh(
\mathbf W_s^{\mathrm{den}}
\mathbf d_{t,p}^{\ell,k}
)
\right],\\
\mathbf R_{t,p}^{\ell,k+1}
&=
\mathbf R_{t,p}^{\ell,k}
\operatorname{Rot6DToSO3}
\!\left(
\mathbf e_{6D}
+
\beta_k\alpha_R^{\mathrm{den}}
\tanh(
\mathbf W_R^{\mathrm{den}}
\mathbf d_{t,p}^{\ell,k}
)
\right),
\end{aligned}
\label{eq:app_refinement_update}
\end{equation}
where $\mathbf e_{6D}$ is the identity-centered 6D rotation code,
$\mathbf e_{6D}=[1,0,0,0,1,0]^\top$. The refined cuboid is
\[
\widehat B_{t,p}^{\ell}
=
\bigl(
\mathbf c_{t,p}^{\ell,K_{\mathrm{den}}},
\mathbf s_{t,p}^{\ell,K_{\mathrm{den}}},
\mathbf R_{t,p}^{\ell,K_{\mathrm{den}}}
\bigr).
\]
 
The refiner also predicts quality residuals
$\delta\kappa_{t,p}^{\ell,k}$. The final quality logit and probability are
\begin{equation}
\kappa_{t,p}^{\ell}
=
\kappa_{t,p}^{\ell,0}
+
\frac{
\lambda_{\kappa}^{\mathrm{den}}
}{
K_{\mathrm{den}}
}
\sum_{k=0}^{K_{\mathrm{den}}-1}
\delta\kappa_{t,p}^{\ell,k},
\qquad
q_{t,p}^{\ell}
=
\sigma(
\kappa_{t,p}^{\ell}
).
\label{eq:app_quality_probability}
\end{equation}
 
\subsection{Query--Tube--Memory Refinement}
\label{app:temporal_reasoning_details}
 
\paragraph{Cuboid-quality target.}
For a positive cuboid prediction, define
\begin{equation}
d_c
=
\frac{
\|\widehat{\mathbf c}-\mathbf c\|_2
}{
\|\mathbf s\|_2+\epsilon
},
\qquad
d_s
=
\frac{1}{3}
\|
\log\widehat{\mathbf s}
-
\log\mathbf s
\|_1,
\qquad
d_R
=
\frac{
d_{\mathrm{SO(3)}}
(
\widehat{\mathbf R},\mathbf R
)
}{
\pi
}.
\label{eq:app_quality_components}
\end{equation}
The quality target follows the benchmark's annotated rotation convention
because it is used to rank predictions against the annotated target. ST-D9O
separately marginalizes local-axis flips in its geometry discrepancy. The
detached target combines oriented overlap and parameter agreement:
\begin{equation}
q_{t,p}^{*}
=
\operatorname{stopgrad}\!\left[\eta\,
\operatorname{softIoU}_{4^3}
(
\widehat B_{t,p},B_t
)
+
(1-\eta)
\exp(
-d_c-d_s-d_R
)\right].
\label{eq:app_quality_target}
\end{equation}
Here, $\operatorname{stopgrad}$ detaches the target from the prediction.
We use $\eta=0.6$ and occupancy temperature $0.15$. Valid negative
predictions receive target zero. The quality head is trained with
\begin{equation}
\mathcal L_{\mathrm{quality}}
=
w(q^*,q)
\operatorname{BCE}
(
q,q^*
),
\qquad
w(q^*,q)
=
\begin{cases}
q^*, & q^*>0,\\
\alpha q^\gamma, & q^*=0,
\end{cases}
\label{eq:app_quality_loss}
\end{equation}
where $(\alpha,\gamma)=(0.75,2)$. Quality ranking therefore uses the annotated rotation convention, while
ST-D9O uses the flip-minimized geometry discrepancy.
 
\paragraph{Reference selection.}
Proposal-level ranking combines query association with predicted geometry
quality,
\begin{equation}
\widetilde z_{t,p}^{\ell}
=
z_{t,p}^{\ell}
+
\lambda_q\log(q_{t,p}^{\ell}+\epsilon),
\qquad
p_t^{\star,\ell}
=
\arg\max_{p\in\{1,\ldots,P\}}
\widetilde z_{t,p}^{\ell}.
\label{eq:app_proposal_reliability}
\end{equation}
The absence-adjusted frame response is
\begin{equation}
\nu_t^\ell
=
r_t^\ell
-
\lambda_b\operatorname{softplus}(b_t^\ell).
\label{eq:app_reference_presence_logit}
\end{equation}
The reference-frame score mixes the strongest proposal and the frame
response,
\begin{equation}
\rho_t^\ell
=
\alpha_{\mathrm{mix}}
\widetilde z_{t,p_t^{\star,\ell}}^{\ell}
+
(1-\alpha_{\mathrm{mix}})\nu_t^\ell,
\qquad
\alpha_{\mathrm{mix}}=0.5.
\label{eq:app_frame_reliability}
\end{equation}
The three highest-scoring frames are retained as candidates. Candidates with
$\sigma(\rho_t^\ell)<0.7$ are discarded, while the strongest frame is always
kept. The selected proposal in a retained frame is $p_t^{\star,\ell}$.
We write the retained indices as
$\mathcal S^\ell=\{(t_k^\ell,p_k^\ell)\}_{k=1}^{K_\ell}$, where
$p_k^\ell=p_{t_k^\ell}^{\star,\ell}$ and $1\leq K_\ell\leq3$.
References can come from any occurrence in the search sequence; selecting
the latest occurrence is deferred to final response decoding.
 
\paragraph{Box-local memory.}
For a proposal center $\mathbf p$ and selected cuboid
$B=(\mathbf c,\mathbf s,\mathbf R)$, define
\begin{equation}
d_B(
\mathbf p,B
)
=
\left\|
\operatorname{diag}(\mathbf s/2)^{-1}
\mathbf R^\top
(
\mathbf p-\mathbf c
)
\right\|_2^2.
\label{eq:app_box_distance}
\end{equation}
The local weights and memory token are defined over the query-conditioned
proposal tokens $\mathbf u_{t,p}^{\ell}$ from Eq.~\ref{eq:app_prediction_token}:
\begin{equation}
\begin{aligned}
\alpha_{k,p}^{\ell}
&=
\operatorname{softmax}_{p}
\left[
-d_B
(
\mathbf p_{t_k^\ell,p},
\widehat B_{t_k^\ell,p_k^\ell}^{\ell}
)
\right],\\
\mathbf m_k^\ell
&=
\sum_p
\alpha_{k,p}^{\ell}
\mathbf u_{t_k^\ell,p}^{\ell}.
\end{aligned}
\label{eq:app_box_memory}
\end{equation}
The pooling sums over all proposal tokens in the selected reference frame. The pooled tokens are
augmented with selected geometry in their reference frames
$\mathcal F_{t_k^\ell}^s$, normalized frame position, and reference confidence.
We denote this augmented memory by $\mathcal M^\ell$, distinguishing it from
the pooled descriptors $\mathbf m_k^\ell$ in Eq.~\ref{eq:app_box_memory}.
 
\paragraph{Query and tube updates.}
The Query Updater applies self-attention to the current query tokens,
cross-attention to $\mathcal M^\ell$, and a gated residual anchored to the
original $Q^0$. This retains information from the independent query while
incorporating search observations. The Tube Updater cross-attends proposal
features to the same memory. Its write gate uses box-local Gaussian
compatibility, frame-index temporal decay, reference confidence, and a
validity mask. Both updaters contain two attention layers with three heads.
 
Using the same $\operatorname{Predict}$ interface as the main paper, one
QTM round follows
\begin{equation}
\begin{aligned}
\mathcal H^\ell
&=\operatorname{Predict}(V^\ell,Q^\ell),\\
\mathcal M^\ell
&=\Phi_{\mathrm{mem}}(V^\ell,Q^\ell,\mathcal H^\ell,\boldsymbol\rho^\ell),\\
Q^{\ell+1}
&=\operatorname{QueryUpdater}(Q^\ell,\mathcal M^\ell,Q^0),\\
V^{\ell+1}
&=\operatorname{TubeUpdater}(V^\ell,\mathcal M^\ell),\\
\mathcal H^{\ell+1}
&=\operatorname{Predict}(V^{\ell+1},Q^{\ell+1}).
\end{aligned}
\label{eq:app_query_tube_update}
\end{equation}
Here, $\Phi_{\mathrm{mem}}$ comprises reference selection, pooling of the
query-conditioned proposal tokens, and memory augmentation. Proposal
centers and the query scale remain fixed across rounds; voting, proposal
sampling, and initial local aggregation run once. Current predictions
determine which observations are stored and how their evidence is pooled.
QTM edits no predicted cuboid directly.
The shared fusion, prediction, and within-head box-refinement modules then
produce the next cuboids and scores. Two rounds yield
$(Q^0,V^0,\mathcal H^0)\rightarrow(Q^1,V^1,\mathcal H^1)
\rightarrow(Q^2,V^2,\mathcal H^2)$.
 
\subsection{GT-Near Warm-Up Proposals}
\label{app:gt_near_warmup}
 
During the first $8{,}000$ training iterations, eight GT-centered proposals
with deterministic box-local offsets are appended to each positive frame.
The offsets are scaled by $0.15$ relative to the target cuboid. The number of
warm-up proposals then linearly decays to zero over the following $4{,}000$
iterations. These auxiliary proposals supply nearby regression examples
while voting is being learned; inference uses only the 96 native proposals.
 
\subsection{ST-D9O}
\label{app:std9o_details}
 
ST-D9O distinguishes among the geometry operator, the discrepancy defined on
its outputs, and the predictions to which that discrepancy is applied. The
operator is formulated independently of the PGL-3D inference architecture
and is evaluated only during training.
 
\paragraph{Cuboid geometry.}
For a cuboid
$B=(\mathbf c,\mathbf s,\mathbf R)$ and canonical coordinate
$\mathbf u\in[-1,1]^3$, define
\begin{equation}
\mathbf x_B(\mathbf u)
=
\mathbf c
+
\mathbf R
\operatorname{diag}(\mathbf s/2)
\mathbf u.
\label{eq:app_cuboid_mapping}
\end{equation}
The eight canonical vertices produce the corner set $\mathcal C(B)$.
A $5\times5$ grid on each canonical face produces the surface set
$\mathcal S(B)$, and a $6^3$ canonical grid produces the volume samples
$\mathcal X(B)$.
 
For point $\mathbf x$, define
\begin{equation}
\mathbf d_B(\mathbf x)
=
|
\mathbf R^\top
(
\mathbf x-\mathbf c
)
|
-
\mathbf s/2.
\label{eq:app_box_residual}
\end{equation}
The oriented-box signed distance and soft occupancy are
\begin{equation}
\begin{aligned}
\operatorname{SDF}_B(\mathbf x)
&=
\|
\max(
\mathbf d_B(\mathbf x),0
)
\|_2
+
\min\!\left(
\max_j d_{B,j}(\mathbf x),0
\right),\\
\operatorname{Occ}_B(\mathbf x)
&=
\sigma\!\left(
-\frac{
\operatorname{SDF}_B(\mathbf x)
}{
\tau_{\mathrm{occ}}
}
\right).
\end{aligned}
\label{eq:app_sdf_occupancy}
\end{equation}
The geometry operator is
\begin{equation}
\Gamma(B)
=
(
\mathcal C(B),
\mathcal S(B),
\operatorname{SDF}_B,
\operatorname{Occ}_B
).
\label{eq:app_geometry_operator}
\end{equation}
The mapping is compatible with automatic differentiation and is differentiable
almost everywhere with respect to the cuboid center, positive side lengths,
and continuous rotation representation. Non-smooth branch boundaries arise
from absolute-value, maximum/minimum, clipping, and symmetry-selection
operations, and the soft overlap below is a sampled approximation of 9-DoF
IoU. Exact 9-DoF IoU is used only for evaluation.
 
\paragraph{Geometry discrepancy.}
Given a prediction $\widehat B$ and target $B$, the discrepancy compares the
parametric cuboid and the boundary, field, and overlap representations induced
by $\Gamma$. We first define the parameter and boundary terms.
 
\paragraph{Parameter and boundary terms.}
Let
$\widehat B=(\widehat{\mathbf c},
\widehat{\mathbf s},\widehat{\mathbf R})$ be a prediction. We use
\begin{equation}
\begin{aligned}
\mathcal L_c
&=
\operatorname{SL1}_{0.1}
\left(
\frac{
\widehat{\mathbf c}-\mathbf c
}{
\mathbf s
},
\mathbf 0
\right),\\
\mathcal L_s
&=
\operatorname{SL1}_{0.05}
(
\log\widehat{\mathbf s},
\log\mathbf s
),\\
\mathcal L_v
&=
\operatorname{SL1}_{0.05}
(
\log\operatorname{vol}(\widehat B),
\log\operatorname{vol}(B)
).
\end{aligned}
\label{eq:app_parameter_losses}
\end{equation}
The geodesic rotation distance is
\begin{equation}
d_{\mathrm{SO(3)}}
(
\mathbf R_1,\mathbf R_2
)
=
\arccos\!\left[
\operatorname{clip}
\left(
\frac{
\operatorname{tr}
(
\mathbf R_1^\top\mathbf R_2
)-1
}{2},
-1,1
\right)
\right].
\label{eq:app_rotation_geodesic}
\end{equation}
 
To handle annotation-equivalent local-axis flips, we use
$\mathcal G_{\mathrm{flip}}
=\{\mathbf I,\mathbf R_x(\pi),\mathbf R_y(\pi),\mathbf R_z(\pi)\}$.
For $\mathbf S\in\mathcal G_{\mathrm{flip}}$, let
$B_{\mathbf S}=(\mathbf c,\mathbf s,\mathbf R\mathbf S)$. The symmetry-aware
rotation and boundary terms are
\begin{equation}
\begin{aligned}
\mathcal L_R
&=
\min_{\mathbf S\in\mathcal G_{\mathrm{flip}}}
d_{\mathrm{SO(3)}}
(
\widehat{\mathbf R},
\mathbf R\mathbf S
),\\
\mathcal L_{\mathrm{cor}}
&=
\min_{\mathbf S\in\mathcal G_{\mathrm{flip}}}
\operatorname{SL1}_{0.1}
(
\mathcal C(\widehat B),
\mathcal C(B_{\mathbf S})
),\\
\mathcal L_{\mathrm{surf}}
&=
\min_{\mathbf S\in\mathcal G_{\mathrm{flip}}}
\operatorname{SL1}_{0.1}
(
\mathcal S(\widehat B),
\mathcal S(B_{\mathbf S})
).
\end{aligned}
\label{eq:app_boundary_losses}
\end{equation}
 
\paragraph{Signed-distance term.}
The two directional terms are
\begin{equation}
\begin{aligned}
\mathcal L_{\widehat B\rightarrow B}^{\mathrm{sdf}}
&=
\frac{1}{|\mathcal S(\widehat B)|}
\sum_{\mathbf x\in\mathcal S(\widehat B)}
\operatorname{SL1}_{\beta_{\mathrm{sdf}}}
\left(
\frac{
\operatorname{SDF}_B(\mathbf x)
}{
\|\mathbf s\|_2+\epsilon
},
0
\right),\\
\mathcal L_{B\rightarrow\widehat B}^{\mathrm{sdf}}
&=
\frac{1}{|\mathcal S(B)|}
\sum_{\mathbf x\in\mathcal S(B)}
\operatorname{SL1}_{\beta_{\mathrm{sdf}}}
\left(
\frac{
\operatorname{SDF}_{\widehat B}(\mathbf x)
}{
\|\widehat{\mathbf s}\|_2+\epsilon
},
0
\right).
\end{aligned}
\label{eq:app_directional_sdf}
\end{equation}
We set
$\mathcal L_{\mathrm{sdf}}
=\tfrac12(
\mathcal L_{\widehat B\rightarrow B}^{\mathrm{sdf}}
+
\mathcal L_{B\rightarrow\widehat B}^{\mathrm{sdf}})$ and
$\beta_{\mathrm{sdf}}=0.05$.
 
\paragraph{Soft overlap.}
The directional coverage estimates are
\begin{equation}
\begin{aligned}
C_{\widehat B\leftarrow B}
&=
\frac{1}{|\mathcal X(B)|}
\sum_{\mathbf x\in\mathcal X(B)}
\operatorname{clip}_{[0,1]}
\frac{
\operatorname{Occ}_{\widehat B}(\mathbf x)
}{
\operatorname{Occ}_B(\mathbf x)+\epsilon
},\\
C_{B\leftarrow\widehat B}
&=
\frac{1}{|\mathcal X(\widehat B)|}
\sum_{\mathbf x\in\mathcal X(\widehat B)}
\operatorname{clip}_{[0,1]}
\frac{
\operatorname{Occ}_B(\mathbf x)
}{
\operatorname{Occ}_{\widehat B}(\mathbf x)+\epsilon
}.
\end{aligned}
\label{eq:app_directional_coverage}
\end{equation}
Let $V_B=\operatorname{vol}(B)$ and
$V_{\widehat B}=\operatorname{vol}(\widehat B)$. We define
\begin{equation}
I_{\mathrm{soft}}
=
\min\!\left\{
\frac{
C_{\widehat B\leftarrow B}V_B
+
C_{B\leftarrow\widehat B}V_{\widehat B}
}{2},
V_B,
V_{\widehat B}
\right\},
\label{eq:app_soft_intersection}
\end{equation}
and
\begin{equation}
\operatorname{softIoU}
(
\widehat B,B
)
=
\frac{
I_{\mathrm{soft}}
}{
V_{\widehat B}
+
V_B
-
I_{\mathrm{soft}}
+
\epsilon
}.
\label{eq:app_soft_iou}
\end{equation}
 
The per-cuboid discrepancy of Eq.~\ref{eq:gst-discrepancy} is
\begin{equation}
\begin{aligned}
D_{\mathrm{9D}}(\widehat B,B)
={}&
\lambda_c\mathcal L_c
+
\lambda_s\mathcal L_s
+
\lambda_v\mathcal L_v
+
\lambda_R\mathcal L_R\\
&+
\lambda_{\mathrm{cor}}\mathcal L_{\mathrm{cor}}
+
\lambda_{\mathrm{surf}}\mathcal L_{\mathrm{surf}}
+
\lambda_{\mathrm{sdf}}\mathcal L_{\mathrm{sdf}}\\
&+
\lambda_{\mathrm{iou}}
[
1-\operatorname{softIoU}(\widehat B,B)
].
\end{aligned}
\label{eq:app_std9o_total}
\end{equation}
Thus, $D_{\mathrm{param}}=\lambda_c\mathcal L_c+
\lambda_s\mathcal L_s+\lambda_v\mathcal L_v+\lambda_R\mathcal L_R$,
$D_{\mathrm{boundary}}=\lambda_{\mathrm{cor}}\mathcal L_{\mathrm{cor}}+
\lambda_{\mathrm{surf}}\mathcal L_{\mathrm{surf}}$,
$D_{\mathrm{SDF}}=\lambda_{\mathrm{sdf}}\mathcal L_{\mathrm{sdf}}$, and
$D_{\mathrm{overlap}}=\lambda_{\mathrm{iou}}[1-\operatorname{softIoU}]$.
The weights in these groups are applied once.
 
\paragraph{Selected and dense supervision.}
$\mathcal L_{\mathrm{selected}}^\ell$ applies $D_{\mathrm{9D}}$ to the most
reliable proposal prediction in each positive frame.
$\mathcal L_{\mathrm{dense}}^\ell$ applies it to predictions whose centers
are within $0.3$ m of the target center or are among the eight nearest
proposal centers. During this phase, the dense set may also include GT-near
warm-up proposals. ST-D9O supervises the decoded cuboids, and the geometry
gradients propagate through the cuboid prediction heads into the proposal
features. Its temporal application is defined separately below for contiguous
positive runs.
 
\subsection{Temporal Geometry Regularization}
\label{app:temporal_regularization}
 
Let $\widetilde B_t^\ell$ be the selected cuboid in positive frame $t$.
Temporal regularization is applied only within contiguous positive response
runs. The predictions remain in their native camera coordinates
$\mathcal F_t^s$, so this term is a low-weight smoothness prior on adjacent
predictions in the same sensor frame. It combines center
acceleration, adjacent log-size change, $\mathrm{SO}(3)$ rotation change, and
corner displacement:
\begin{equation}
\begin{aligned}
\mathcal L_{\mathrm{temp}}^\ell
={}&
\lambda_{\mathrm{motion}}
\mathcal L_{\mathrm{motion}}^\ell
+
\lambda_{\mathrm{size}}
\mathcal L_{\mathrm{size}}^\ell\\
&+
\lambda_{\mathrm{rot}}
\mathcal L_{\mathrm{rot}}^\ell
+
\lambda_{\mathrm{corner}}
\mathcal L_{\mathrm{corner}}^\ell.
\end{aligned}
\label{eq:app_temporal_total}
\end{equation}
The component definitions are
\begin{equation}
\begin{aligned}
\mathcal L_{\mathrm{motion}}^\ell
&=
\sum_t
\operatorname{SL1}_{0.05}
(
\widetilde{\mathbf c}_t^\ell
-
2\widetilde{\mathbf c}_{t-1}^\ell
+
\widetilde{\mathbf c}_{t-2}^\ell,
\mathbf 0
),\\
\mathcal L_{\mathrm{size}}^\ell
&=
\sum_t
\operatorname{SL1}_{0.05}
(
\log\widetilde{\mathbf s}_t^\ell
-
\log\widetilde{\mathbf s}_{t-1}^\ell,
\mathbf 0
),\\
\mathcal L_{\mathrm{rot}}^\ell
&=
\sum_t
d_{\mathrm{SO(3)}}
(
\widetilde{\mathbf R}_t^\ell,
\widetilde{\mathbf R}_{t-1}^\ell
),\\
\mathcal L_{\mathrm{corner}}^\ell
&=
\sum_t
\operatorname{SL1}_{0.1}
(
\mathcal C(\widetilde B_t^\ell)
-
\mathcal C(\widetilde B_{t-1}^\ell),
\mathbf 0
).
\end{aligned}
\label{eq:app_temporal_components}
\end{equation}
These adjacent-parameter differences act as weak regularization; they do
not model target dynamics or compensate camera ego-motion. Pair and triplet
masks include only consecutive positive frames; negative and ignored frames
break the masks. We use
$(\lambda_{\mathrm{motion}},
\lambda_{\mathrm{size}},
\lambda_{\mathrm{rot}},
\lambda_{\mathrm{corner}})
=(1.0,0.2,0.2,0.5)$.
 
\subsection{Training Objective}
\label{app:training_details}
 
Point voting and proposal classification use balanced focal losses. Proposal
ranking combines nearest-positive cross-entropy, a hard-negative margin on
positive frames, and Top-8 suppression on valid negative frames. Frame
response uses class-balanced BCE. Cuboid quality uses
Eq.~\ref{eq:app_quality_loss}, and target absence uses BCE with the
presence--absence margin.
 
The initialization objective is
\begin{equation}
\mathcal L_{\mathrm{init}}
=
2.0\mathcal L_{\mathrm{pt\mbox{-}obj}}
+
0.5\mathcal L_{\mathrm{vote}}.
\label{eq:app_initialization_loss}
\end{equation}
To make the grouped objective in Eq.~\ref{eq:gst-objective} explicit, define
\begin{equation}
\begin{aligned}
\mathcal L_{\mathrm{disc}}^\ell
&=2.0\mathcal L_{\mathrm{prop\mbox{-}cls}}^\ell
+\mathcal L_{\mathrm{rank}}^\ell
+2.0\mathcal L_{\mathrm{resp}}^\ell
+0.5\mathcal L_{\mathrm{abs}}^\ell,
\end{aligned}
\label{eq:app_grouped_losses}
\end{equation}
and let $\mathcal L_{\mathrm{selected}}^\ell$ and $\mathcal L_{\mathrm{dense}}^\ell$
average $D_{\mathrm{9D}}$ over the selected and the dense positive
predictions of stage $\ell$ (Appendix~\ref{app:std9o_details}).
$\mathcal L_{\mathrm{quality}}^\ell$ denotes the loss in
Eq.~\ref{eq:app_quality_loss} before its outer coefficient of $0.5$.
At stage $\ell$, the fully expanded objective is
\begin{equation}
\begin{aligned}
\mathcal L_{\mathrm{stage}}^\ell
={}&
2.0\mathcal L_{\mathrm{prop\mbox{-}cls}}^\ell
+
1.0\mathcal L_{\mathrm{rank}}^\ell
+
2.0\mathcal L_{\mathrm{resp}}^\ell\\
&+
0.5\mathcal L_{\mathrm{quality}}^\ell
+
0.5\mathcal L_{\mathrm{abs}}^\ell
+
1.0\mathcal L_{\mathrm{selected}}^\ell\\
&+
0.25\mathcal L_{\mathrm{dense}}^\ell
+
0.05\mathcal L_{\mathrm{temp}}^\ell.
\end{aligned}
\label{eq:app_stage_loss}
\end{equation}
All three prediction stages are supervised, so the same geometry operator
shapes both the initial proposal features and the features produced by the two
QTM updates:
\begin{equation}
\mathcal L
=
\mathcal L_{\mathrm{init}}
+
\frac{
3\mathcal L_{\mathrm{stage}}^0
+
2\mathcal L_{\mathrm{stage}}^1
+
\mathcal L_{\mathrm{stage}}^2
}{6}.
\label{eq:app_multistage_loss}
\end{equation}
The fixed-step refiner has no separate objective; gradients reach it through
the quality loss, the selected and dense geometry terms, and the temporal
regularization.
 
\subsection{Response-Tube Decoding}
\label{app:inference_details}
 
During inference, GT-near warm-up proposals and ST-D9O are disabled. Cuboid
quality is used internally for memory selection, whereas the raw association
logits serve as unary scores for selecting one cuboid candidate per frame.
 
For each frame, the inference wrapper retains the eight cuboid predictions
with the highest $z_{t,p}^{2}$. The center, size, and rotation terms defined
below are low-weight continuity penalties between adjacent cuboids in the
camera coordinates of each frame. They encode no registered object motion. For path
$\boldsymbol p=(p_1,\ldots,p_T)$, they are
\begin{equation}
\begin{aligned}
\mathcal C_t^c
&=
\frac{
\|
\widehat{\mathbf c}_{t,p_t}
-
\widehat{\mathbf c}_{t-1,p_{t-1}}
\|_2
}{
\frac12
(
\|\widehat{\mathbf s}_{t,p_t}\|_1
+
\|\widehat{\mathbf s}_{t-1,p_{t-1}}\|_1
)
+
\epsilon
},\\
\mathcal C_t^s
&=
\|
\log\widehat{\mathbf s}_{t,p_t}
-
\log\widehat{\mathbf s}_{t-1,p_{t-1}}
\|_2,\\
\mathcal C_t^R
&=
\left\|
\operatorname{wrap}
\left(
\boldsymbol\theta_{t,p_t}
-
\boldsymbol\theta_{t-1,p_{t-1}}
\right)
\right\|_2.
\end{aligned}
\label{eq:app_path_transitions}
\end{equation}
Here $\boldsymbol\theta_{t,p}$ denotes the Euler-angle representation used by
the inference implementation, and $\operatorname{wrap}$ maps each angular
difference to $[-\pi,\pi)$. This low-weight rotation term is used only for
path smoothing; the predicted cuboids and ST-D9O rotation loss remain
full-$\mathrm{SO}(3)$.
 
The path cost is
\begin{equation}
\mathcal J(
\boldsymbol p
)
=
\sum_{t=1}^{T}
-\log\sigma(
z_{t,p_t}^{2}
)
+
\sum_{t=2}^{T}
(
0.2\mathcal C_t^c
+
0.05\mathcal C_t^s
+
0.02\mathcal C_t^R
).
\label{eq:app_path_cost}
\end{equation}
Dynamic programming returns one refined cuboid $\widehat B_t$ per frame.
 
The final response probability is
$s_t=\sigma(\nu_t^2)$. After median filtering, local peaks below
$\eta_{\mathrm{peak}}=0.8$ times the strongest peak are removed. The latest
remaining peak is selected, and the interval expands in both directions while
the smoothed response exceeds $0.7$ times the selected peak. This yields
$\widehat{\mathcal R}$ and
$\widehat{\mathcal T}
=\{(t,\widehat B_t):t\in\widehat{\mathcal R}\}$.
The tube confidence is $\max_{t\in\widehat{\mathcal R}}s_t$.
 
\section{Evaluation Protocol Details}
\label{app:evaluation_protocol}
 
A 3DVQL sequence may contain multiple visible response segments, while
the retrieval target is the most recent contiguous occurrence. For each
query, evaluation uses one Top-1 response tube and its confidence.
 
Temporal average precision (tAP) evaluates interval localization using
temporal IoU. It averages AP over $\{0.25,0.50,0.75,0.95\}$;
$\mathrm{tAP}_\tau$ denotes AP at threshold $\tau$.
 
For spatio-temporal evaluation, frame-wise 3D overlaps between predicted
and target 9-DoF cuboids are aggregated into $\mathrm{stIoU}_{3D}$.
$\mathrm{stAP}_\tau$ denotes AP at threshold $\tau$. The aggregate stAP is
the mean of these values over the evaluator's threshold set
$\mathcal I_{\mathrm{st}}$:
\begin{equation}
\mathrm{stAP}=\frac{1}{|\mathcal I_{\mathrm{st}}|}
\sum_{\tau\in\mathcal I_{\mathrm{st}}}\mathrm{stAP}_\tau.
\label{eq:app_mean_stap}
\end{equation}
We report $\mathrm{stAP}_{0.05}$ and $\mathrm{stAP}_{0.25}$ separately.
 
Success is the percentage of queries satisfying
$\mathrm{stIoU}_{3D}\geq0.05$. Recovery is the percentage of ground-truth
response frames whose predicted cuboid reaches frame-wise 3D IoU of at
least $0.5$. Both are reported in percent. These benchmark metrics are
computed from the Top-1 predictions.
 
\paragraph{Stage-wise diagnostics.}
Table~\ref{tab:gst-stages} reports additional diagnostics for
$\mathcal H^0$, $\mathcal H^1$, and $\mathcal H^2$. For each stage, we decode
a response tube from $\mathcal H^\ell$ with the procedure of
Appendix~\ref{app:inference_details} and compare it with the annotation on
the ground-truth response frames. Frame IoU is the mean frame-wise 3D IoU
between the decoded cuboid and the annotated cuboid over these frames. Center
error is the mean Euclidean distance between centers in meters, size error is
the mean absolute difference of the three side lengths in meters, and
rotation error is the mean flip-minimized geodesic distance of
Eq.~\ref{eq:app_boundary_losses} in degrees. Reference precision is the
percentage of frames selected by Eq.~\ref{eq:gst-reference} at stage $\ell$
that lie inside a ground-truth response interval and whose selected cuboid
reaches a 3D IoU of at least $0.25$ with the annotation. For
$\mathcal H^2$, the selection rule is applied offline, since no further QTM
round follows. These diagnostics measure the effect of each QTM round within
one trained model; they are not a comparison of models trained with different
numbers of rounds.
 
\section{Reproducibility and Baseline Details}
\label{app:implementation}
 
\subsection{Training and Architecture Configuration}
 
All models use the official 3DVQL split. The benchmark has no public
validation split. We therefore report one fixed configuration and the final
training checkpoint. We do not tune the proposal budget, loss weights,
or thresholds on the official test set. All ablations inherit the same
sampling strategy, optimization budget, and inference procedure, and apply
the intervention stated for that analysis
(Table~\ref{tab:app_variant_definitions}).
Tables~\ref{tab:training_configuration},~\ref{tab:architecture_details}, and
~\ref{tab:geometry_inference_details} list the base configuration settings.
 
\begin{table}[H]
\centering
\caption{Training and optimization configuration of PGL-3D.}
\label{tab:training_configuration}
\small
\setlength{\tabcolsep}{4pt}
\begin{tabular*}{\linewidth}{@{\extracolsep{\fill}}lc@{}}
\toprule
Setting & Value \\
\midrule
Training split & 1,601 sequences / 131.4K frames / 5,157 tracklets \\
Training duration & 160K data-loading iterations (about 400 epochs) \\
Optimizer updates & About 80K (gradient accumulation over 2 mini-batches) \\
Training hardware & 4 $\times$ NVIDIA RTX 4090 \\
Batch per GPU / per step / effective update & 1 / 4 / 8 \\
Optimizer & AdamW \\
Base learning rate & $1\times10^{-4}$ \\
Weight decay & $5\times10^{-3}$ \\
Mixed precision & Disabled (FP32) \\
Gradient clipping & Maximum norm $5.0$ \\
Base configuration seed (control for all ablations) & 42 \\
Complete-model repeat seeds & 42, 43, 44 \\
Model selection & Final training snapshot \\
\bottomrule
\end{tabular*}
\end{table}
 
\begin{table}[H]
\centering
\caption{Architecture and refinement settings of PGL-3D.}
\label{tab:architecture_details}
\small
\setlength{\tabcolsep}{4pt}

\begin{tabularx}{\linewidth}{
>{\raggedright\arraybackslash}X
>{\centering\arraybackslash}X
}
\toprule
\rowcolor{gray!12}
Setting & Value \\
\midrule
Sampled video length $T$ & 30 \\
Raw points per frame & 3,072 \\
Search / query candidates & 192 / 96 \\
Query tokens $K_q$ & 32 \\
Proposal candidates per frame $P$ & 96 \\
QTM rounds / prediction stages & 2 / 3 ($\ell=0,1,2$) \\
Fixed refinement steps $K_{\mathrm{den}}$ & 2 \\
Temporal radius / point neighbors & 5 frames / 8 per neighboring frame \\
RGB / temporal residual strengths & 0.25 / 0.35 \\
Current-frame local aggregation & 3 query-scale-normalized neighborhoods \\
Local candidates per neighborhood & 32 \\
Center / log-size decode scales & $(4.0,0.35)$ \\
Initial rotation decoder & Absolute Rot6D in the current observation \\
Refinement scales $(\alpha_c^{\mathrm{den}},\alpha_s^{\mathrm{den}},\alpha_R^{\mathrm{den}})$ & $(0.35,0.12,0.25)$ \\
Refinement quality-residual weight & 0.3 \\
Top reference candidates / threshold & 3 / 0.7 \\
Reference weights $(\lambda_q,\lambda_b,\epsilon)$ in Eq.~\ref{eq:gst-reference} & {0.7 / 0.8 / \ensuremath{10^{-4}}} \\
Shared token dimension & 96 \\
Query Updater layers / heads & 2 / 3 \\
Tube Updater layers / heads & 2 / 3 \\
\bottomrule
\end{tabularx}
\end{table}
 
\begin{table}[H]
\centering
\caption{Geometry, warm-up, and inference settings of PGL-3D.}
\label{tab:geometry_inference_details}
\small
\setlength{\tabcolsep}{4pt}

\begin{tabularx}{\linewidth}{
>{\raggedright\arraybackslash}X
>{\centering\arraybackslash}X
}
\toprule
\rowcolor{gray!12}
Setting & Value \\
\midrule
Quality grid / occupancy temperature & $4^3$ / 0.15 \\
Quality IoU / parameter weights & 0.6 / 0.4 \\
Varifocal $(\alpha,\gamma)$ & $(0.75,2)$ \\
ST-D9O surface / volume sampling & 6 faces $\times$ $5\times5$ / $6^3$ \\
ST-D9O occupancy temperature & 0.1 \\
ST-D9O weights (Eq.~\ref{eq:app_std9o_total} order)
& $(1.0,0.5,0.1,0.5,1.0,0.25,0.5,0.5)$ \\
Dense-positive radius / nearest count & 0.3 m / 8 \\
GT-near proposals / local-offset scale & 8 / 0.15 \\
Warm-up full phase / decay & first 8K / next 4K iterations \\
Refinement perturbation start / warm-up & 12K / 4K iterations \\
Maximum perturbation & $0.03\,\bar s^q$ / 0.05 / 0.10 rad \\
Inference candidates per frame & Top-8 association logits \\
Path weights & $(0.2,0.05,0.02)$ \\
Peak-retention / boundary ratios & 0.8 / 0.7 \\
Response decoder & Latest valid peak with bidirectional expansion \\
\bottomrule
\end{tabularx}
\end{table}
 
The ST-D9O tuple is ordered as
$(\lambda_c,\lambda_s,\lambda_v,\lambda_R,\lambda_{\mathrm{cor}},
\lambda_{\mathrm{surf}},\lambda_{\mathrm{sdf}},\lambda_{\mathrm{iou}})$.
The complete-model repeats use seeds 42, 43, and 44 and are reported in
Table~\ref{tab:app_seed_results}. Component and stage-wise analyses use the seed-42 run
(0.777 tAP, 0.266 stAP) as their control. Because no public validation split
is available, all variants use the fixed configuration and the final
checkpoint.
 
\subsection{Baseline Reproduction and PRVQL-3D Adaptation}
\label{app:baseline_reproduction}
 
AF, GAF, PAF, and LaF are quoted from the original 3DVQL
comparison~\citep{laf}. PRVQL-3D preserves the progressive query--video
refinement principle of PRVQL~\citep{prvql}, uses RGB--point-cloud search
input, and replaces the original 2D response and box outputs with a frame-wise
9-DoF cuboid prediction interface. It is evaluated with the official 3DVQL
split and evaluator used by PGL-3D.
 
\begin{table}[H]
\centering
\caption{Scope of the main 3DVQL comparisons. ``Reported'' rows are quoted
from the benchmark paper; PRVQL-3D shares the official input, split, and
evaluator and differs from PGL-3D in architecture.}
\label{tab:app_baseline_scope}
\footnotesize
\setlength{\tabcolsep}{2.4pt}
\renewcommand{\arraystretch}{1.0}
\begin{tabularx}{\linewidth}{@{}>{\raggedright\arraybackslash}p{0.24\linewidth}
>{\raggedright\arraybackslash}p{0.18\linewidth}
>{\centering\arraybackslash}p{0.10\linewidth}
>{\centering\arraybackslash}p{0.12\linewidth}
>{\raggedright\arraybackslash}X@{}}
\toprule
\rowcolor{gray!12}
Method group & Result source & RGB--PC & Official eval. & PGL-3D-specific modules \\
\midrule
AF / GAF / PAF / LaF & Reported~\citep{laf} & Yes & Yes & None \\
PRVQL-3D & Our adaptation & Yes & Yes & No QTM or ST-D9O \\
PGL-3D & Our implementation & Yes & Yes & Full model \\
\bottomrule
\end{tabularx}
\end{table}
 
Table~\ref{tab:app_baseline_scope} summarizes the scope of each comparison.
The main table establishes system-level performance, and the component
analyses in Sec.~\ref{sec:ablation_study} attribute gains to intermediate
geometry, QTM, and ST-D9O within PGL-3D.
 
Under this protocol, PRVQL-3D obtains 0.546 tAP, 0.685
tAP$_{0.25}$, 0.099 mean stAP, 0.402 stAP$_{0.05}$, 0.025
stAP$_{0.25}$, 2.00 Recovery, and 55.4 Success. It serves as a stronger
response-refinement reference under the official split and evaluator. Its gap
to PGL-3D in Table~\ref{tab:gst-main} therefore includes architectural
differences and is not attributed to any single component.
 
\section{Additional Experimental Analysis}
\label{app:additional_experiments}
 
The following analyses complement the main comparisons with complete-model
repeats, supervision and feedback variants, detailed component ablations,
geometry-objective comparisons, and complete transfer and efficiency results.
 
\subsection{Complete-Model Results Across Seeds}
\label{app:seed_results}
 
Table~\ref{tab:app_seed_results} gives the three runs underlying the main
comparison, followed by their summary statistics. The seed-42 run is the
control of all component analyses. Ablation differences smaller than the
seed-level spread reported here are not resolved by single runs.
 
\begin{table}[H]
\centering
\caption{Complete-model performance across seeds. Rec. and Succ. are
percentages. The last two rows report the mean and standard deviation.}
\label{tab:app_seed_results}

\footnotesize
\setlength{\tabcolsep}{2pt}

\begin{tabularx}{\linewidth}{
l
*{7}{>{\raggedleft\arraybackslash}X}
}
\toprule
\rowcolor{gray!12}
Seed & tAP & $\mathrm{tAP}_{0.25}$ & stAP & $\mathrm{stAP}_{0.05}$ &
$\mathrm{stAP}_{0.25}$ & Rec.\% & Succ.\% \\
\midrule

42 & 0.777 & 0.850 & 0.266 & 0.729 & 0.215 & 6.66 & 79.3 \\
43 & 0.752 & 0.828 & 0.270 & 0.745 & 0.203 & 5.72 & 78.1 \\
44 & 0.767 & 0.830 & 0.274 & 0.763 & 0.222 & 4.72 & 79.1 \\

\midrule

\rowcolor{pglhighlight}
\textbf{Mean}
& \textbf{0.766}
& \textbf{0.836}
& \textbf{0.270}
& \textbf{0.746}
& \textbf{0.213}
& \textbf{5.70}
& \textbf{78.8} \\

Std.
& 0.013
& 0.012
& 0.004
& 0.017
& 0.010
& 0.97
& 0.66 \\

\bottomrule
\end{tabularx}
\end{table}

\subsection{Supervision and Geometry-Feedback Ablations}
\label{app:feedback_ablations}
 
Table~\ref{tab:app_feedback_complete} expands
Table~\ref{tab:gst-feedback} to all seven reported metrics, and
Table~\ref{tab:app_variant_definitions} defines every ablation variant of
the paper by the switches it changes. Without cuboid-conditioned pooling,
the memory token of a selected reference frame is the query-conditioned
token of its selected proposal, and the updaters are otherwise unchanged.

\begin{table}[H]
\centering
\caption{Definition of the ablation variants by the switches they change.
Stages: number of prediction stages; supervised: stages that receive the
objective of Eq.~\ref{eq:app_stage_loss}; pooling: cuboid-conditioned soft
pooling of Eq.~\ref{eq:gst-memory}; signals: terms of
Eq.~\ref{eq:gst-reference} used for reference selection; Q-Upd.\ and T-Upd.:
Query Updater and Tube Updater of Eq.~\ref{eq:gst-update}.
$^\ast$Tube Updater without its temporal-decay gate.}
\label{tab:app_variant_definitions}

\footnotesize
\setlength{\tabcolsep}{1.5pt}
\renewcommand{\arraystretch}{1.08}

\begin{tabularx}{\linewidth}{
>{\raggedright\arraybackslash}X
cccccc
}
\toprule
\rowcolor{gray!12}
Variant & Stages & Supervised & Pooling & Signals & Q-Upd. & T-Upd. \\
\midrule

Full PGL-3D
& 3 & $\ell=0,1,2$ & \cmark & $z,q,r,b$ & \cmark & \cmark \\

Final-stage supervision only (Table~\ref{tab:gst-feedback})
& 3 & $\ell=2$ & \xmark & $z$ & \cmark & \cmark \\

Intermediate supervision (Table~\ref{tab:gst-feedback})
& 3 & $\ell=0,1,2$ & \xmark & $z$ & \cmark & \cmark \\

Association-only QTM (Table~\ref{tab:gst-feedback})
& 3 & $\ell=0,1,2$ & \cmark & $z$ & \cmark & \cmark \\

Final prediction only (Table~\ref{tab:app_complete_ablations})
& 1 & final & \xmark & -- & \cmark & \cmark \\

Initial stage only (Table~\ref{tab:app_qtm_update_paths})
& 1 & $\ell=0$ & -- & -- & \xmark & \xmark \\

Query Updater only (Table~\ref{tab:app_qtm_update_paths})
& 3 & $\ell=0,1,2$ & \cmark & $z,q,r,b$ & \cmark & \xmark \\

Tube Updater only (Table~\ref{tab:app_qtm_update_paths})
& 3 & $\ell=0,1,2$ & \cmark & $z,q,r,b$ & \xmark & \cmark \\

w/o cuboid quality (Table~\ref{tab:gst-components})
& 3 & $\ell=0,1,2$ & \cmark & $z,r,b$ & \cmark & \cmark \\

w/o Target Absence (Table~\ref{tab:gst-components})
& 3 & $\ell=0,1,2$ & \cmark & $z,q,r$ & \cmark & \cmark \\

w/o Temporal Feature Refinement (Table~\ref{tab:gst-components})
& 3 & $\ell=0,1,2$ & \cmark & $z,q,r,b$ & \cmark & \cmark$^\ast$ \\

\bottomrule
\end{tabularx}
\end{table}

\begin{table}[H]
\centering
\caption{Complete supervision and QTM-guidance ablations. Rec. and Succ.
are percentages; the full-model row is the seed-42 control.}
\label{tab:app_feedback_complete}

\footnotesize
\setlength{\tabcolsep}{1.5pt}
\renewcommand{\arraystretch}{1.08}

\begin{tabularx}{\linewidth}{
>{\raggedright\arraybackslash}X
rrrrrrr
}
\toprule
\rowcolor{gray!12}
Variant & tAP & $\mathrm{tAP}_{0.25}$ & stAP &
$\mathrm{stAP}_{0.05}$ & $\mathrm{stAP}_{0.25}$ &
Rec.\% & Succ.\% \\
\midrule

Final-stage supervision only
& 0.702 & 0.769 & 0.225 & 0.650 & 0.158 & 4.73 & 72.2 \\

Intermediate supervision
& 0.721 & 0.792 & 0.246 & 0.697 & 0.179 & 5.28 & 74.9 \\

Association-only QTM
& 0.745 & 0.821 & 0.254 & 0.719 & 0.188 & 5.92 & 77.6 \\

\rowcolor{pglhighlight}
\textbf{Full PGL-3D}
& \textbf{0.777}
& \textbf{0.850}
& \textbf{0.266}
& \textbf{0.729}
& \textbf{0.215}
& \textbf{6.66}
& \textbf{79.3} \\

\bottomrule
\end{tabularx}
\end{table}

\subsection{Complete Main-Ablation Results}
\label{app:complete_ablations}
 
Table~\ref{tab:app_complete_ablations} provides complete results for the
component ablations and geometry-objective comparison, including the
experiments summarized in Table~\ref{tab:gst-components} and the training
warm-up analysis. Component variants apply the named intervention to the
seed-42 configuration; Table~\ref{tab:app_variant_definitions} lists the
switches of each variant, and the MGIoU variant is discussed in
Appendix~\ref{app:std9o_incremental}.

The results separate temporal retrieval from cuboid localization.
Removing RGB or cuboid quality raises tAP but lowers
$\mathrm{stAP}_{0.25}$, so these components' benefits are not uniform
across metrics. Temporal Feature Aggregation, Target Absence, and Temporal
Feature Refinement improve strict localization in their respective
comparisons. Separate Regression gives 0.183 stAP and 0.076
$\mathrm{stAP}_{0.25}$, compared with 0.266 and 0.215 for the full model.
The MGIoU variant obtains 0.220 and 0.160 for these two metrics,
respectively. Removing GT-near proposals from the full model reduces
stAP and $\mathrm{stAP}_{0.25}$ to 0.248 and 0.187.

\begin{table}[H]
\centering
\caption{Complete results for the main component ablations and
geometry-objective comparison. Rec. and Succ. are reported in percent.
The full-model rows repeat the seed-42 control.}
\label{tab:app_complete_ablations}

\footnotesize
\setlength{\tabcolsep}{1.2pt}
\renewcommand{\arraystretch}{1.08}

\begin{tabularx}{\linewidth}{
>{\raggedright\arraybackslash}p{0.38\linewidth}
*{7}{>{\raggedleft\arraybackslash}X}
}
\toprule
\rowcolor{gray!12}
Variant
& tAP
& $\mathrm{tAP}_{0.25}$
& stAP
& $\mathrm{stAP}_{0.05}$
& $\mathrm{stAP}_{0.25}$
& Rec.
& Succ. \\
\midrule

\multicolumn{8}{c}{\textbf{(a) Multimodal Encoding}} \\

\rowcolor{pglhighlight}
\textbf{Full PGL-3D}
& \textbf{0.777}
& \textbf{0.850}
& \textbf{0.266}
& \textbf{0.729}
& \textbf{0.215}
& \textbf{6.66}
& \textbf{79.3} \\

w/o RGB
& 0.803 & 0.852 & 0.235 & 0.689 & 0.181 & 5.66 & 75.1 \\

w/o Temporal Feature Aggregation
& 0.735 & 0.798 & 0.232 & 0.669 & 0.176 & 5.34 & 74.0 \\

\midrule

\multicolumn{8}{c}{\textbf{(b) Intermediate 9-DoF Prediction}} \\

\rowcolor{pglhighlight}
\textbf{Full PGL-3D}
& \textbf{0.777}
& \textbf{0.850}
& \textbf{0.266}
& \textbf{0.729}
& \textbf{0.215}
& \textbf{6.66}
& \textbf{79.3} \\

Final prediction only
& 0.698 & 0.762 & 0.209 & 0.621 & 0.151 & 4.47 & 70.6 \\

\midrule

\multicolumn{8}{c}{\textbf{(c) QTM Refinement}} \\

\rowcolor{pglhighlight}
\textbf{Full PGL-3D}
& \textbf{0.777}
& \textbf{0.850}
& \textbf{0.266}
& \textbf{0.729}
& \textbf{0.215}
& \textbf{6.66}
& \textbf{79.3} \\

w/o cuboid quality
& 0.831 & 0.902 & 0.246 & 0.735 & 0.154 & 5.82 & 78.8 \\

w/o Target Absence
& 0.741 & 0.806 & 0.254 & 0.685 & 0.202 & 6.08 & 75.9 \\

w/o Temporal Feature Refinement
& 0.728 & 0.791 & 0.227 & 0.666 & 0.174 & 5.21 & 73.6 \\

\midrule

\multicolumn{8}{c}{\textbf{(d) ST-D9O}} \\

\rowcolor{pglhighlight}
\textbf{Full PGL-3D}
& \textbf{0.777}
& \textbf{0.850}
& \textbf{0.266}
& \textbf{0.729}
& \textbf{0.215}
& \textbf{6.66}
& \textbf{79.3} \\

Separate Regression
& 0.817 & 0.863 & 0.183 & 0.672 & 0.076 & 3.03 & 73.3 \\

w/ MGIoU
& 0.760 & 0.840 & 0.220 & 0.680 & 0.160 & 5.00 & 76.0 \\

w/o Dense Loss
& 0.763 & 0.829 & 0.251 & 0.706 & 0.190 & 5.97 & 76.9 \\

w/o Temporal Loss
& 0.770 & 0.837 & 0.258 & 0.712 & 0.199 & 6.18 & 77.6 \\

\midrule

\multicolumn{8}{c}{\textbf{(e) Training Warm-Up}} \\

\rowcolor{pglhighlight}
\textbf{Full PGL-3D}
& \textbf{0.777}
& \textbf{0.850}
& \textbf{0.266}
& \textbf{0.729}
& \textbf{0.215}
& \textbf{6.66}
& \textbf{79.3} \\

w/o GT-near proposals
& 0.758 & 0.823 & 0.248 & 0.697 & 0.187 & 5.88 & 76.4 \\

\bottomrule
\end{tabularx}
\end{table}

\subsection{Additional Analysis of Multimodal Encoding}
\label{app:temporal_cue_dependence}
 
Table~\ref{tab:app_spatiotemporal_cues} separates the descriptors used by
Temporal Feature Aggregation. These are one-factor diagnostics of the fixed
configuration.
 

\begin{table}[H]
\centering
\caption{Ablation of the descriptors used by Temporal Feature Aggregation.}
\label{tab:app_spatiotemporal_cues}

\footnotesize
\setlength{\tabcolsep}{2pt}
\renewcommand{\arraystretch}{1.08}

\begingroup
\renewcommand{\tabularxcolumn}[1]{m{#1}}

\begin{tabularx}{\linewidth}{
>{\raggedright\arraybackslash}m{0.21\linewidth}
>{\raggedright\arraybackslash}X
*{6}{>{\centering\arraybackslash}m{0.066\linewidth}}
}
\toprule
\rowcolor{gray!12}
Variant
& Temporal descriptor
& tAP
& $\mathrm{tAP}_{0.25}$
& stAP
& $\mathrm{stAP}_{0.05}$
& $\mathrm{stAP}_{0.25}$
& Succ.\% \\
\midrule

w/o Temporal Feature Aggregation
& None
& 0.735 & 0.798 & 0.232 & 0.669 & 0.176 & 74.0 \\

Feature and time only
& Feature difference + time offset
& 0.763 & 0.836 & 0.247 & 0.706 & 0.194 & 77.0 \\

w/o relative xyz / distance
& Feature + time + previous displacement
& 0.769 & 0.841 & 0.254 & 0.716 & 0.202 & 77.8 \\

w/o previous displacement
& Feature + relative coordinates + time
& 0.773 & 0.845 & 0.260 & 0.724 & 0.207 & 78.6 \\

\rowcolor{pglhighlight}
\textbf{Full}
& Feature + coordinates + time + displacement
& \textbf{0.777}
& \textbf{0.850}
& \textbf{0.266}
& \textbf{0.729}
& \textbf{0.215}
& \textbf{79.3} \\

\bottomrule
\end{tabularx}

\endgroup
\end{table}

Feature and time already improve stAP$_{0.25}$ from 0.176 to 0.194. Relative
coordinates and previous-frame displacement add smaller gains. An improvement over frame-wise evidence therefore remains when these
coordinate terms are removed. The coordinate terms therefore act as auxiliary observation features.
 
\subsection{Additional Analysis of QTM Update Paths}
\label{app:qtm_update_paths}
 
Table~\ref{tab:app_qtm_update_paths} separates the Query Updater and Tube
Updater. The updater-only and full variants use the same three-stage prediction
schedule. ``Initial stage only'' is trained without QTM, so its 0.178 stAP differs
from the 0.229 stAP of $\mathcal H^0$ in Table~\ref{tab:gst-stages}, which
is the first-stage output of the full model.
 
\begin{table}[H]
\centering
\caption{Ablation of the two QTM update paths. Rec. and Succ. are reported in
percent.}
\label{tab:app_qtm_update_paths}

\footnotesize
\setlength{\tabcolsep}{1.2pt}
\renewcommand{\arraystretch}{1.08}

\begin{tabularx}{\linewidth}{
>{\raggedright\arraybackslash}X
cc
rrrrrrr
}
\toprule
\rowcolor{gray!12}
Variant
& Q-updater
& Tube-updater
& tAP
& $\mathrm{tAP}_{0.25}$
& stAP
& $\mathrm{stAP}_{0.05}$
& $\mathrm{stAP}_{0.25}$
& Rec.
& Succ. \\
\midrule

Initial stage only
& No & No
& 0.704 & 0.782 & 0.178 & 0.612 & 0.128 & 4.22 & 71.4 \\

Tube Updater only
& No & Yes
& 0.742 & 0.816 & 0.221 & 0.684 & 0.171 & 5.24 & 75.6 \\

Query Updater only
& Yes & No
& 0.751 & 0.823 & 0.232 & 0.695 & 0.181 & 5.47 & 76.4 \\

\rowcolor{pglhighlight}
\textbf{Full QTM}
& \textbf{Yes}
& \textbf{Yes}
& \textbf{0.777}
& \textbf{0.850}
& \textbf{0.266}
& \textbf{0.729}
& \textbf{0.215}
& \textbf{6.66}
& \textbf{79.3} \\

\bottomrule
\end{tabularx}
\end{table}
 
Both update paths improve the initial-stage model. The Query Updater provides
the larger individual gain, while their combination gives the best temporal
and 3D localization results. This analysis directly supports the use of both
updates in QTM.
 
 

\subsection{Additional Analysis of ST-D9O}
\label{app:std9o_incremental}
 
\begin{table}[H]
\centering
\caption{Incremental analysis of ST-D9O. Each row adds one level of geometry
supervision while keeping the remaining model and training settings fixed.}
\label{tab:app_std9o_incremental}

\footnotesize
\setlength{\tabcolsep}{1.5pt}
\renewcommand{\arraystretch}{1.06}

\begin{tabularx}{\linewidth}{
>{\raggedright\arraybackslash}p{0.18\linewidth}
>{\raggedright\arraybackslash}X
*{6}{c}
}
\toprule
\rowcolor{gray!12}
Variant
& Added supervision
& tAP
& $\mathrm{tAP}_{0.25}$
& stAP
& $\mathrm{stAP}_{0.05}$
& $\mathrm{stAP}_{0.25}$
& Succ.\% \\
\midrule

Separate Regression
& Center, size, rotation
& \textbf{0.817}
& \textbf{0.863}
& 0.183
& 0.672
& 0.076
& 73.3 \\

Parameter terms
& Normalized center, log-size/volume, rotation
& 0.801
& 0.856
& 0.211
& 0.688
& 0.118
& 74.9 \\

+ boundary terms
& Corners and surfaces
& 0.790
& 0.852
& 0.231
& 0.704
& 0.151
& 76.2 \\

+ SDF
& Signed-distance geometry
& 0.781
& 0.849
& 0.244
& 0.714
& 0.169
& 77.1 \\

+ Soft Occupancy
& Soft-overlap term
& 0.775
& 0.846
& 0.256
& 0.723
& 0.196
& 78.2 \\

+ dense positives
& Dense positive predictions
& 0.774
& 0.847
& 0.261
& 0.726
& 0.207
& 78.8 \\

\rowcolor{pglhighlight}
\textbf{Full ST-D9O}
& Temporal regularization
& 0.777
& 0.850
& \textbf{0.266}
& \textbf{0.729}
& \textbf{0.215}
& \textbf{79.3} \\

\bottomrule
\end{tabularx}
\end{table}

Strict stAP$_{0.25}$ increases from 0.076 with Separate Regression to 0.215
with the full objective. Parameter, boundary, SDF, and Soft Occupancy terms
provide successive gains. Dense positive supervision and temporal
regularization add further improvements. Each level of supervision adds to the previous one, which supports
combining these terms into one objective. The individual terms have
precedents in prior work (Sec.~\ref{sec:related_work}); the contribution of
ST-D9O lies in their combination and stage-wise application. The ``+ dense positives'' row and
the ``w/o Temporal Loss'' row of Table~\ref{tab:app_complete_ablations} are
separate runs of the same configuration; their difference (0.003 stAP) is
within the seed-level spread of Table~\ref{tab:app_seed_results}.
 
\paragraph{Alternative geometry objective.}
Table~\ref{tab:app_complete_ablations} additionally reports the MGIoU
variant, which replaces the soft-overlap term $D_{\mathrm{overlap}}$ of
ST-D9O with the MGIoU loss of \citet{mgiou} and keeps the remaining terms.
It obtains 0.220 stAP and 0.160 $\mathrm{stAP}_{0.25}$,
compared with 0.266 and 0.215 for full PGL-3D with ST-D9O.
The absolute gains are 0.046 and 0.055, corresponding to relative
improvements of 20.9\% and 34.4\%, respectively. Recovery increases
from 5.00\% to 6.66\%, and Success from 76.0\% to 79.3\%.
 
This comparison complements the incremental analysis in
Table~\ref{tab:app_std9o_incremental}. MGIoU compares the marginal extents
of the two cuboids along their face normals, whereas the soft overlap of
ST-D9O samples occupancy inside both cuboids and is paired with boundary
and signed-distance terms that respond to partial misalignment.
 
 
\subsection{Additional Analysis of Response-Tube Decoding}
\label{app:response_decoding_analysis}
 
\begin{table}[H]
\centering
\caption{Effects of the temporal training term and path-transition costs. Each
row removes one component from the fixed configuration.}
\label{tab:app_temporal_heuristics}

\begin{tabularx}{\linewidth}{
>{\hspace{-\tabcolsep}\raggedright\arraybackslash}X
rrr
}
\toprule
\rowcolor{gray!12}
Variant
& stAP
& $\mathrm{stAP}_{0.05}$
& $\mathrm{stAP}_{0.25}$ \\
\midrule

\rowcolor{pglhighlight}
\textbf{Full PGL-3D}
& \textbf{0.266}
& \textbf{0.729}
& \textbf{0.215} \\

w/o Temporal Loss
& 0.258
& 0.712
& 0.199 \\

Unary-only 3D Box Path Decoding
& 0.247
& 0.710
& 0.189 \\

\bottomrule
\end{tabularx}
\end{table}
The frame-local ST-D9O terms remain effective when the Temporal Loss is
removed. The unary-only decoder also retains most of the improvement over
Separate Regression. The temporal training term and the path-transition costs therefore add to a
gain that comes mainly from the frame-local terms.
 
\subsection{Generalization to 3D Tracking}
\label{app:gsot_protocol}
 
Table~\ref{tab:app_gsot_full} places the controlled comparison of
Sec.~\ref{sec:gst_transfer} among the published GSOT3D baselines. Our
reproduction of PROT3D obtains 21.63 mAO, 19.38 mSR$_{50}$, and 5.08
mSR$_{75}$, close to the published 21.97, 19.76, and 5.22~\citep{gsot3d}.
Starting from this reproduction, we keep the backbone, progressive tracking
architecture, training data, optimizer schedule, and inference procedure
fixed and replace only the training-time geometry objective with ST-D9O.
 

\begin{table}[H]
\centering
\caption{Complete comparison on the GSOT3D point-cloud benchmark. Published
baseline results are reported by \citet{gsot3d}. The controlled objective
replacement compares the last two rows. All metrics are percentages.}
\label{tab:app_gsot_full}

\setlength{\tabcolsep}{6pt}

\begin{tabularx}{\linewidth}{
>{\hspace{-\tabcolsep}\raggedright\arraybackslash}X
rrr
}
\toprule
\rowcolor{gray!12}
Method
& mAO
& $\mathrm{mSR}_{50}$
& $\mathrm{mSR}_{75}$ \\
\midrule

P2B
& 9.79
& 8.59
& 1.75 \\

BAT
& 6.56
& 3.54
& 0.88 \\

PTTR
& 14.00
& 10.42
& 1.60 \\

M2-Track
& 20.26
& 14.34
& 1.88 \\

CXTrack
& 14.29
& 8.39
& 1.02 \\

MBPTrack
& 20.54
& 16.55
& 2.57 \\

SeqTrack-3D
& 8.61
& 5.25
& 1.11 \\

M3SOT
& 17.40
& 12.47
& 1.74 \\

PROT3D (reported)
& 21.97
& 19.76
& 5.22 \\

\midrule

PROT3D (our reproduction)
& 21.63
& 19.38
& 5.08 \\

\rowcolor{pglhighlight}
\textbf{PROT3D + ST-D9O}
& \textbf{25.78}
& \textbf{23.61}
& \textbf{6.23} \\

\bottomrule
\end{tabularx}
\end{table}
 
\subsection{Inference Efficiency}
\label{app:efficiency}
 
Table~\ref{tab:app_efficiency} reports the three-method timing comparison.
Measurements use a single RTX~4090 GPU and batch size 1, with cached
DINOv2 features and disk I/O excluded. The three methods process different numbers of frames, and the timings
exclude RGB feature extraction, so they describe the evaluated
configurations. PGL-3D uses 9.8\,GB peak inference memory. ST-D9O is evaluated
only during training and adds no inference-time computation.
 




\begin{table}[H]
\centering
\caption{Inference efficiency under the stated cached-feature protocol.
Input frame counts differ between methods.}
\label{tab:app_efficiency}

\begin{tabularx}{\linewidth}{
>{\hspace{-\tabcolsep}\raggedright\arraybackslash}X
rrr
}
\toprule
\rowcolor{gray!12}
Method
& Frames
& Time (s)
& FPS \\
\midrule

LaF
& 20
& 17.4
& 1.15 \\

PRVQL-3D
& 40
& 2.06
& 19.4 \\

\rowcolor{pglhighlight}
\textbf{PGL-3D}
& \textbf{30}
& \textbf{1.99}
& \textbf{15.1} \\

\bottomrule
\end{tabularx}
\end{table}

\section{Limitations}
\label{app:limitations}
 
PGL-3D prevents direct transfer of the independently collected query pose to
search-side localization. Its query representation nevertheless depends on
the annotated query cuboid and its local-axis convention. The fixed axis-flip
set handles generic $180^\circ$ equivalences but does not enumerate axis
permutations for near-cubic objects or continuous category-specific
symmetries. In addition, the current cuboid-quality target uses the annotated
rotation convention directly rather than the flip-marginalized rotation used
by ST-D9O, so an annotation-equivalent flipped prediction can receive a lower
quality target.
 
3DVQL defines frame-wise boxes in calibrated camera coordinates, and PGL-3D
does not estimate inter-frame alignment or ego-motion. The cross-frame
coordinate descriptors, the temporal geometry term, and the decoder
continuity costs therefore operate in the sensor frame and do not model
registered motion. The analyses in
Appendices~\ref{app:temporal_cue_dependence} and
\ref{app:response_decoding_analysis} examine these components under the
camera motion of the benchmark; robustness to arbitrary camera motion or
calibration drift remains untested.
The current evaluation covers 3DVQL and transfer to one generic 3D tracker on
GSOT3D. Broader validation on full-rotation detection, additional tracking
architectures, and substantially different sensor configurations remains
future work.
 
\FloatBarrier
 
\section{Additional Qualitative Results}
\label{app:qualitative}
 
The first visualization compares initial cuboids with predictions after two
QTM rounds; the second summarizes representative successes and failures.
QTM usually improves size and orientation consistency, while severe occlusion,
extreme point sparsity, and geometrically similar distractors remain the main
failure modes.
 
\begin{figure}[H]
\centering
\vspace{15pt}
\begin{subfigure}[c]{\linewidth}
\centering
\includegraphics[width=\linewidth]{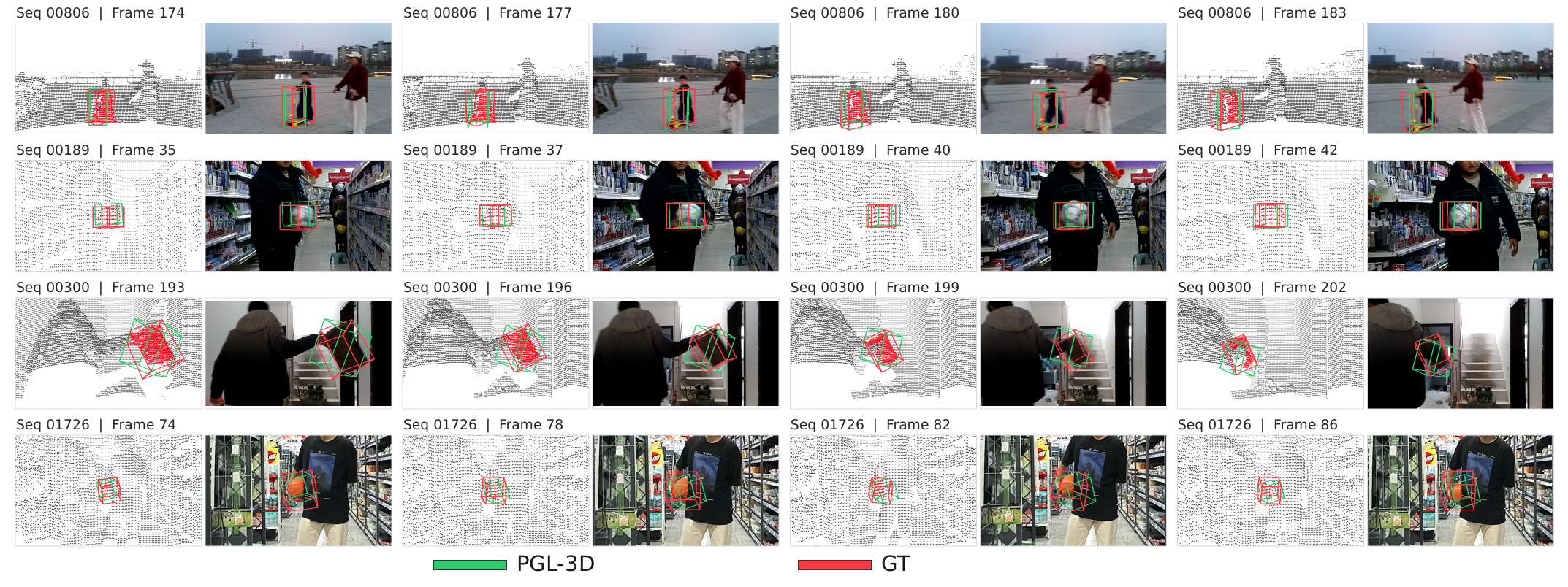}
\caption{Initial predictions and results after two QTM rounds.}
\label{fig:appendix_tube_refinement}
\end{subfigure}

\vspace{15pt}

\begin{subfigure}[c]{\linewidth}
\centering
\includegraphics[width=\linewidth]{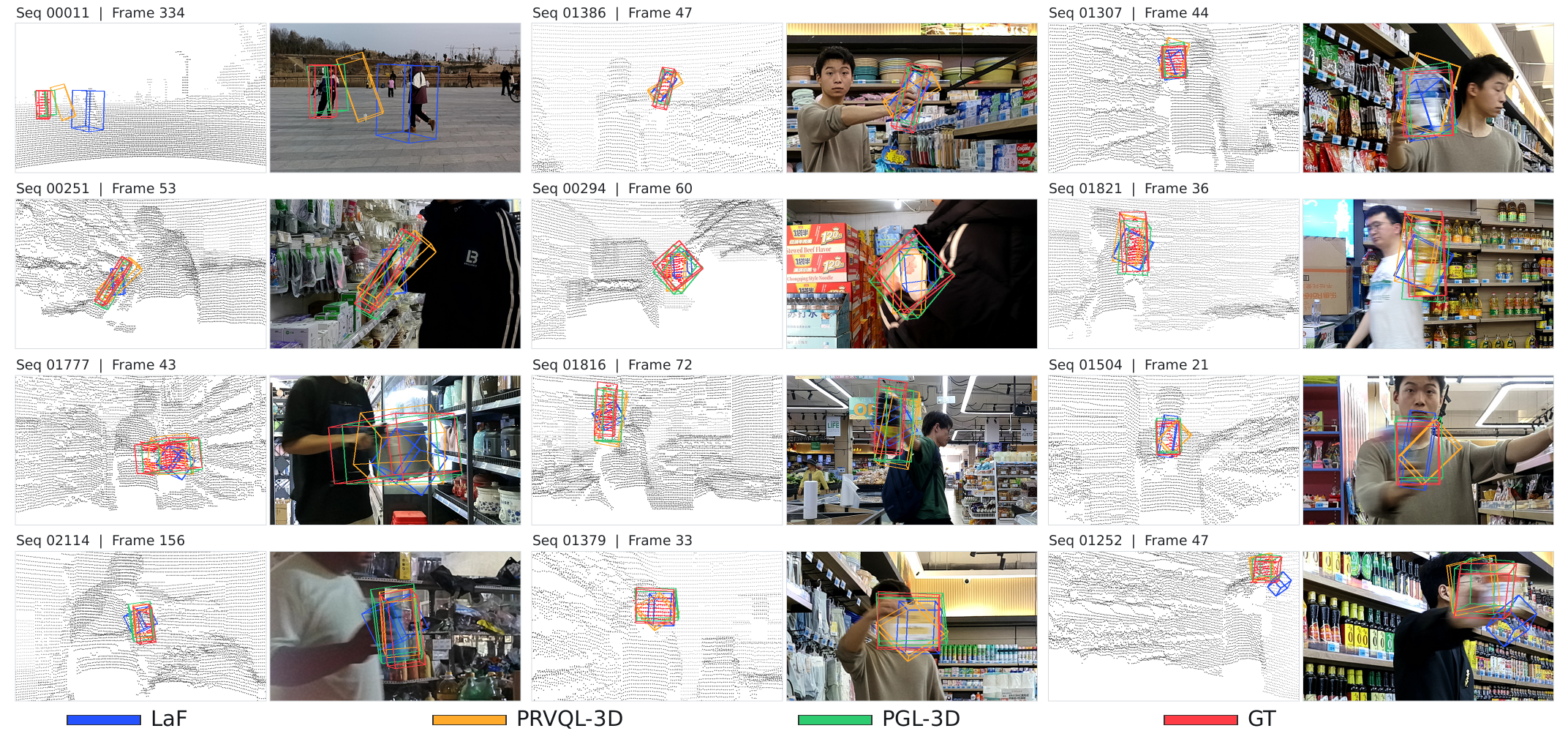}
\caption{Additional success and failure cases.}
\label{fig:appendix_success_failure}
\end{subfigure}

\caption{Additional qualitative results. Green and red denote PGL-3D and
ground-truth cuboids.}
\label{fig:appendix_qualitative}
\end{figure}

\end{document}

%% file: math_commands.tex
\usepackage{amsmath,amsfonts,bm}

\def\eqref#1{equation~\ref{#1}}

\def\1{\bm{1}}

\DeclareMathAlphabet{\mathsfit}{\encodingdefault}{\sfdefault}{m}{sl}
\SetMathAlphabet{\mathsfit}{bold}{\encodingdefault}{\sfdefault}{bx}{n}

